%% file: MachineLearningForMFG-arxiv.tex
\documentclass[9pt,legal,twocolumn]{scrartcl}
\usepackage{amssymb,amsfonts,amsmath,latexsym,dsfont}
\usepackage{fullpage}
\usepackage{graphicx}
\usepackage{mathrsfs}
 \usepackage[margin=16mm]{geometry}
 \usepackage[T1]{fontenc}

\input{abbrv.tex}

 \let\OLDthebibliography\thebibliography
 \renewcommand\thebibliography[1]{
   \OLDthebibliography{#1}
   \setlength{\parskip}{0pt}
   \setlength{\itemsep}{0pt plus 0.3ex}
 }

 \usepackage[font=small,labelfont=bf]{caption}
 \usepackage{authblk}
 
 \newcommand{\dropcap}[1]{#1}
\usepackage{hyperref}
\hypersetup{
 citecolor=blue,
 urlcolor=blue,
 pdftitle={A Machine Learning Framework for Solving High-Dimensional Mean Field Game and Mean Field Control Problems},
 pdfauthor={Lars Ruthotto, Stanley Osher, Wuchen, Li, Levon Nurbekyan, Samy Wu Fung},
 pdfkeywords={mean field games, mean field control, machine learning, optimal control, neural networks, optimal transport, Hamilton-Jacobi-Bellman equations}
 }

\graphicspath{{Figures/}{./}}

\title{A Machine Learning Framework for Solving High-Dimensional Mean Field Game and Mean Field Control Problems}
\author[a]{Lars Ruthotto}
\author[b]{Stanley Osher} 
\author[b]{Wuchen Li} 
\author[b]{Levon Nurbekyan} 
\author[b]{Samy Wu Fung} 

\affil[a]{Department of Mathematics,  Emory University, Atlanta, GA, USA (\url{lruthotto@emory.edu})}
\affil[b]{Department of Mathematics,  University of California, Los Angeles, CA, USA}

\begin{document}
\maketitle
\begin{abstract}
	\textbf{Abstract.}
	\input{abstract.tex}
\end{abstract}

\bigskip


\bigskip
\section*{Introduction}
\input{document.tex}

\section*{Acknowledgements}
\input{acknowledgements.tex}

\bibliographystyle{abbrv}
\bibliography{main,MFGML}

\newpage
\normalfont
\newpage
\input{appendix.tex}
\end{document}

%% file: abbrv.tex
\newcommand{\hf}{{\frac 12}}

\newcommand{\bfI}{{\bf I}}

\newcommand{\bfe}{{\bf e}}

\newcommand{\bfm}{{\bf m}}

\newcommand{\CO}{{\cal O}}
\newcommand{\CP}{{\cal P}}

\newcommand{\CF}{{\cal F}}
\newcommand{\CG}{{\cal G}}

\newcommand{\bfSigma}{{\boldsymbol \Sigma}}

\newcommand{\R}{\ensuremath{\mathds{R}}}

%% file: abstract.tex
Mean field games (MFG) and mean field control (MFC) are critical classes of multi-agent models for efficient analysis of massive populations of interacting agents. 
Their areas of application span topics in economics, finance, game theory, industrial engineering, crowd motion, and more. 
In this paper, we provide a flexible machine learning framework for the numerical solution of potential MFG and MFC models. 
State-of-the-art numerical methods for solving such problems utilize spatial discretization that leads to a curse-of-dimensionality. 
We approximately solve high-dimensional problems by combining  Lagrangian and Eulerian viewpoints and leveraging recent advances from machine learning. 
More precisely, we work with a Lagrangian formulation of the problem and enforce the underlying Hamilton-Jacobi-Bellman (HJB) equation that is derived from the Eulerian formulation.
Finally,  a tailored neural network parameterization of the MFG/MFC solution helps us avoid any spatial discretization. 
Our numerical results include the approximate solution of 100-dimensional instances of optimal transport and crowd motion problems on a standard work station and a validation using a Eulerian solver in two dimensions.
These results open the door to much-anticipated applications of MFG and MFC models that were beyond reach with existing numerical methods.

%% file: document.tex
\dropcap{M}ean field games (MFG) \cite{LasryLions06a,LasryLions06b,LasryLions2007,HCM06,HCM07} and mean field control (MFC) \cite{bensoussan2013} allow one to simulate and analyze interactions within large populations of agents.
Hence, these models have found wide spread use in
economics \cite{gueantlasrylions11,moll14,Gomes:2015th,moll17}, finance \cite{lehalle16,lehalle18,jaimungal19,caines17}, crowd motion \cite{wolfram11,wolfram14,aurell18,achdou19}, industrial engineering \cite{kizikale19,paola19,gomessaude'18},  and more recently in data science \cite{EHanLi2018_meanfield} and material dynamics \cite{welch19}. 

The theoretical properties of MFG and MFC problems have been continuously developed over the last few decades; see, e.g., \cite{carda19,gangbo15,gomes_book16,cirant19,delarue18a,delarue18b}. 
A key observation is that both problems involve a Hamilton-Jacobi-Bellman (HJB) equation that is coupled with a continuity equation. 
From the solution of this system of partial differential equations (PDE), each agent can infer their optimal action without the need for individual optimization \cite{evans10,soner2006}.
Reminiscent of similar conventions in physics, we call the solution of the HJB a potential since its gradient defines the agent's optimal action.

Our framework applies to a common subclass of MFGs, namely potential MFGs, and MFCs.
These problems can be formulated as infinite-dimensional optimal control problems in density space.
Interestingly, their optimality conditions coincide with the HJB and continuity equation. 

Despite many theoretical advances, the development of numerical methods for solving MFGs, particularly in high-dimensional sample spaces, lags and has not kept pace with growing data and problem sizes. 
A crucial disadvantage of most existing approaches for solving MFGs or their underlying HJB equations is their reliance on grids; see ~\cite{achdou13,silva15,achdou16,gomes17,bencarsan'17,bencarmarnen'18,nurbe18,matt'18,Chow2019,silva19,JacobsLegerLiOsher2019_solving} and references therein. 
Grid-based methods are prone to the curse of dimensionality, i.e., their computational complexity grows exponentially with spatial dimension~\cite{Bellman1957}. 

In this paper, we tackle the curse of dimensionality in two steps.
First, extending the approach in~\cite{nursaude18}, we solve the continuity equation and compute all other terms involved in the MFG using Lagrangian coordinates.
In practice, this requires computation of characteristic curves and the Jacobian determinant of the induced transformation; both terms can be inferred from the potential.
In our examples, the former follows the gradient, and the logarithm of the latter can be approximated by integrating the Laplacian of the potential.
Our scheme also allows us to control and enforce the HJB equation along the characteristics. 
These computations can be performed independently and in parallel for all agents.

Second, we parameterize the potential in space and time using a neural network that is specifically designed for an accurate and efficient Lagrangian scheme.
With this design, we can also directly penalize the violation of the HJB equation. 
Thereby, we develop a generic framework that transforms a wide range of MFGs into machine learning (ML) problems. 

In our numerical experiments, we solve high-dimensional instances of the dynamical optimal transport (OT) problem \cite{BenamouBrenier2000,villani2008optimal},  a prototypical instance of a potential MFG, and a mean field game inspired by crowd motion.
In OT, one seeks to find the path connecting two given densities that minimizes a kinetic energy that models transport costs. 
As for other MFGs, there are many relevant high-dimensional instances related to dynamical OT, e.g., in machine learning~\cite{Rezende:2015vu}.
We validate our scheme using comparisons to a Eulerian method on two-dimensional instances.
Most crucially, we show the accuracy and scalability of our method using OT and MFG instances in up to one hundred space dimensions.
We also show that penalizing the HJB violations allows us to solve the problem more accurately with less computational effort.
Our results for the crowd motion problem also show that our framework is capable of solving potential MFGs and MFCs with nonlinear characteristics. 

To enable further theoretical and computational advances as well as applications to real-world problems, we provide our prototype implementation written in Julia~\cite{Bezanson:2017gd} as open-source software under a permissible license.

\section*{Related work}
Our work lies at the interface of machine learning (ML), partial differential equations (PDEs), and optimal control.
Recently, this area has received a lot of attention, rendering a comprehensive review to be beyond the scope of this paper.
The idea of solving high-dimensional PDEs and control problems with neural networks has been pioneered by the works~\cite{E2016,Han:2016ku,HanEtAl2017} and has been further investigated by~\cite{Sirignano:ik}.
In this section, we put our contributions into context by reviewing recent works that combine concepts from machine learning, optimal transport, mean field games, and Lagrangian methods for MFG.

\subsection*{Optimal Transport and ML}
Despite tremendous theoretical insight gained into the problem of optimally transporting one density to match another one, its numerical solution remains challenging, particularly when the densities live in spaces of dimension four or more.
In small-dimensional cases, there are many state-of-the-art approaches that effectively compute the global solution; see, e.g.,\cite{Chow2019,JacobsLegerLiOsher2019_solving, LiRyuOsherYinGangbo2017_parallel, HaberHoresh2015} and the recent survey \cite{PeyreCuturi2018_computationalb}. 
Due to their reliance on Euclidean coordinates, those techniques require spatial discretization, which makes them prone to the curse of dimensionality. 
An exception is the approach in~\cite{Li2018ConstrainedDO} that uses a generative model for computing the optimal transport. 
This work uses a neural network parameterization for the density and a Lagrangian PDE solver.

A machine learning task that bears many similarities with optimal transport is variational inference with normalizing flows~\cite{Rezende:2015vu}.
Roughly speaking, the goal is to transform given samples from a typically unknown distribution such that they are approximately normally distributed.
To this end, one trains a neural network based flow model; hence, the name normalizing flow.
The trained flow can be used as a generative model to produce new samples from the unknown distribution by reversing the flow, starting with samples drawn from a normal distribution.
While the original formulation of the learning problem in normalizing flows does not incorporate transport costs, ~\cite{ZhangEtAl2018,YangEtAl2019,Lin:2019ui} successfully apply concepts from optimal transport to analyze and improve the learning of flows. 
The approach in~\cite{Lin:2019ui} is formulated as a point cloud matching problem and estimates the underlying densities using Gaussian mixture models.
The works~\cite{ZhangEtAl2018,YangEtAl2019} propose neural network models for the potential instead of the velocity of the flow, which leads to more physically plausible models.
This parameterization has also been used to enforce parts of the HJB equation via quadratic penalties~\cite{YangEtAl2019}.
We generalize these ideas from optimal transport to a broader class of mean field games that naturally incorporate transport costs. We also add a penalty for the final time condition of the HJB to the training problem.

\subsection*{Mean Field Games and ML}
Machine learning and MFGs have become increasingly intertwined in recent years.
On the one hand, MFG theory has been used to provide valuable insight into the training problems in deep learning~\cite{EHanLi2018_meanfield}. 
On the other hand,~\cite{YangYeTrivediXuZha2017_learning, CarmonaLauriere2019_convergence, CarmonaLauriere2019_convergencea} use machine learning to solve MFGs in spatial dimensions up to four. 
The methods in \cite{CarmonaLauriere2019_convergence, CarmonaLauriere2019_convergencea} are limited to MFGs whose formulations do not involve the population density, whose computation is challenging.
For the time-independent second-order problems in \cite{CarmonaLauriere2019_convergencea}, one can express the density explicitly in terms of the potential. 
Furthermore, in numerical examples for the time-dependent case in \cite{CarmonaLauriere2019_convergence}, the congestion terms depend on the average positions of the population.
In this situation the congestion term can be computed using sample-averages. 
Our framework does not have the above limitations and, in particular, is applicable to MFGs where there is no analytical formula for the density or special structure that can be used to compute the congestion term, e.g., MFGs with  nonlinear congestion terms.
We achieve this using the Lagrangian formulation that includes an  estimate of the density along the agents' trajectories. 
This generality is a critical contribution of our work.

Additionally, our neural network architecture for the control respects the structure induced by optimality conditions. 
We believe that this property is critical for obtaining accurate algorithms that scale and yield correct geometry for the agents' trajectories. 
As a result, we use our method to approximately solve MFGs in 100 dimensions on a standard work station.

\subsection*{Lagrangian Methods in MFG} 
\label{sub:lagrangian_methods}
To the best of our knowledge, the first Lagrangian method for solving MFG problems appeared in \cite{nursaude18}.  Lagrangian techniques are natural from an optimal control perspective and unavoidable for high-dimensional problems.
However, a crucial computational challenge in applying these techniques in MFG stems from the density estimation, which is critical, e.g., to compute the congestion-cost incurred by an individual agent. 
 In \cite{nursaude18}, the authors overcome this difficulty for non-local interactions by passing to Fourier coordinates in the congestion term and thus avoiding the density estimation. 
 Our neural network parameterization aims to reduce the computational effort and memory footprint of the methods in~\cite{nursaude18} and provides a tractable way to estimate the population density. 

Lagrangian methods for mass-transport problems in image processing were proposed in~\cite{MangRuthotto2017}. 
While the computation of the characteristics is mesh-free, the final density is computed using a particle-in-cell method that does not scale to high-dimensional problems.

\section*{Mathematical Modeling}
\label{sec:modeling}

This section provides a concise mathematical formulation of MFG and MFC models and useful references; for more details, see the monographs~\cite{LasryLions2007,bensoussan2013}.
Mean field games model large populations of rational agents that play the non-cooperative differential game. 
At optimality, this leads to a Nash equilibrium where no single agent can do better by unilaterally changing their strategy. 
By contrast, in mean field control, there is a central planner that aims at a distributed strategy for agents to minimize the average cost or maximize the average payoff across the population. 
Starting with a microscopic description of MFGs, we derive their macroscopic equations, introduce the important class of potential MFGs,  and briefly outline MFCs.

\subsection*{Mean Field Games}

Assume that a continuum population of small rational agents plays a non-cooperative differential game on a time horizon $[0,T]$. Suppose that an agent is positioned at $x \in \mathbb{R}^d$ at time $t\in[0,T]$.
For fixed  $t$, we denote agents' population density by $\rho(\cdot,t) \in \mathcal{P}(\R^d)$, where $\mathcal{P}(\R^d)$ is the space of all probability densities. The agent's cost function  is given by
\begin{equation}\label{eq:single_cost}
    \begin{split}
    J_{x,t}(v,\rho)=\int_t^T L\left(z(s),v(s)\right)+F\left(z(s),\rho(z(s),s)\right)ds
    \\
    + G\left( z(T),\rho(z(T),T) \right),
    \end{split}
\end{equation}
where $v:[0,T]\to \mathbb{R}^d$ is the strategy (control) of this agent, and their position changes according to
\begin{equation}\label{eq:z}
\partial_t z(t)=v\left(t\right),~0\leq t\leq T,\quad z(0)=x.
\end{equation}
In~\eqref{eq:single_cost}, $L: \R^d \times \R^d \to \R$ is a running cost incurred by an agent based solely on their actions, $F:\R^d \times \mathcal{P}(\R^d) \to \R$ is a running cost incurred by an agent based on their interaction with rest of the population, and $G:\R^d \times \mathcal{P}(\R^d) \to \R$ is a terminal cost incurred by an agent based on their final position and the final distribution of the whole population. The terms $F$ and $G$ are called mean field terms because they encode the interaction of a single agent with rest of the population.

The agents forecast a distribution of the population, $\{\rho(\cdot,t)\}_{t=0}^T$, and aim at minimizing their cost. 
Therefore, at a Nash equilibrium, we have that for every $x\in \R^d$
\begin{equation}\label{eq:NE}
J_{x,0}(v,\rho) \leq J_{x,0}(\hat{v},\rho),\quad \forall \hat{v} :[0,T] \to \R^d,
\end{equation}
where $v$ is the equilibrium strategy of an agent at position $x$. 
Here, we assume that agents are small, and their unilateral actions do not alter the density $\rho$.

\begin{figure}
    \begin{center}
        \includegraphics[width=.35\textwidth]{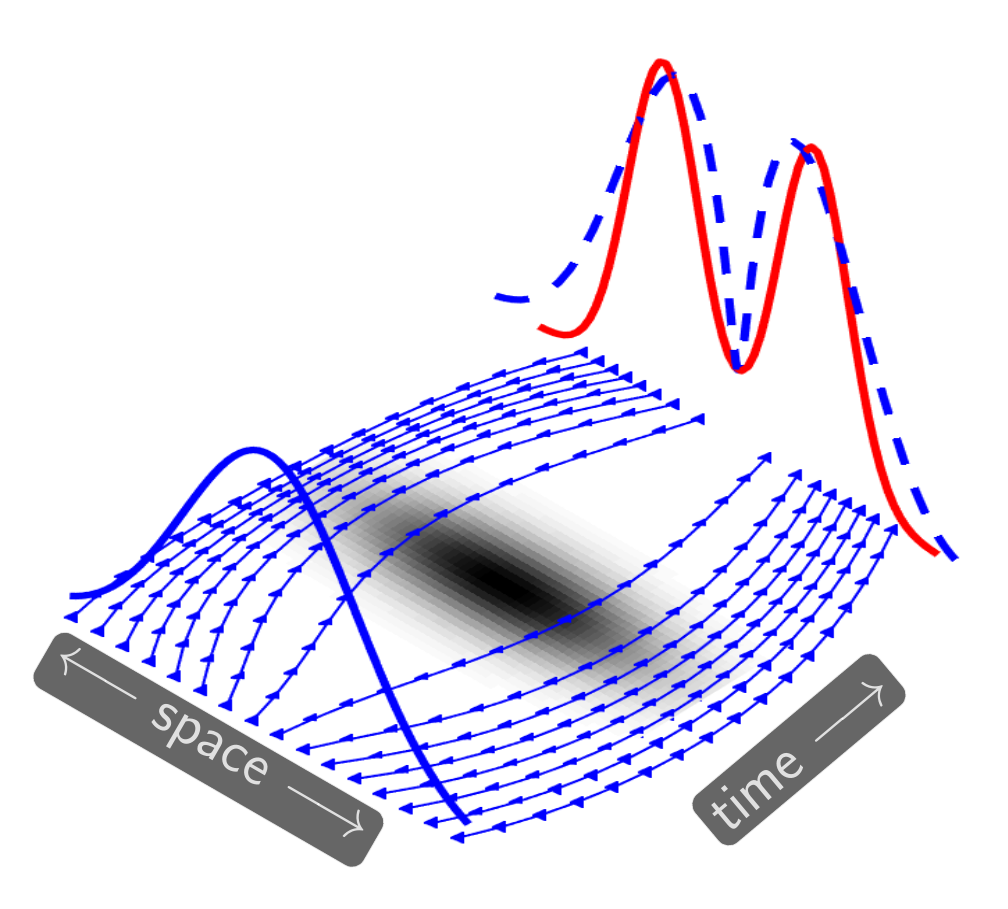}
    \end{center}
    \caption{Illustration of a one-dimensional crowd motion problem. Initially, the crowd of agents is distributed according to $\rho_0$ (thick blue line) and aims at reaching the target distribution $\rho_1$ (red solid line) while avoiding the dark regions in the center of the space-time domain. The blue lines marked with arrows depict the agent's trajectories, i.e., the characteristics. The dashed blue line depicts the push-forward of the initial densities at the final time.   }
    \label{fig1}
\end{figure}

From~\eqref{eq:NE} we have that individual agents solve an optimal control problem that has a value function
\begin{equation}\label{eq:gen_agent_cost}
\begin{split}
\Phi(x,t)=& \inf_{v} J_{x,t}(v,\rho), \text{ s.t.~\eqref{eq:z}}.
\end{split}
\end{equation}
From the optimal control theory (for details see~\cite[Sec.~I.5-6, II.15]{soner2006} or \cite[Sec.~10.3]{evans10}), we have that $\Phi$ solves the Hamilton-Jacobi-Bellmann (HJB) equation 
\begin{equation}\label{eq:HJB}
    \begin{split}
    -\partial_t \Phi(x,t)+H(x,\nabla \Phi(x,t))=& F(x,\rho(x,t)), \\
        \Phi(x,T)=&G(x,\rho(x,T)),
    \end{split}
\end{equation}
where the Hamiltonian, $H:\R^d \times \R^d \to \R$, is defined as
\begin{equation}\label{eq:H}
H(x,p)=\sup_v - p^\top v - L(x,v).
\end{equation}
Furthermore, the Pontryagin Maximum Principle yields that the optimal strategy for an agent at position $x\in\R^d$ and time $t\in(0,T]$ is given by the formula
\begin{equation}\label{eq:v}
v(x,t)=-\nabla_p H(x,\nabla \Phi(x,t) ).
\end{equation} 
Assuming that all agents act optimally according to \eqref{eq:v}, the population density, $\hat{\rho}$, satisfies the continuity equation
\begin{equation}\label{eq:CE}
\begin{split}
\partial_t \hat{\rho}(x,t)-\nabla\cdot \left(\hat{\rho}(x,t) \nabla_p H(x,\nabla \Phi(x,t)) \right)=&0,\\
\hat{\rho}(x,0) =& \rho_0(x),
\end{split}
\end{equation}
where $\rho_0\in \CP(\R^d)$ is the given population density at time $t=0$.
Therefore, an MFG equilibrium is a state when the anticipated distribution coincides with the actual distribution of the population when everyone acts optimally; that is, $\hat{\rho}=\rho$. 

The above discussion suggests an alternative to optimizing the strategy $v$ individually for each agent. 
One can obtain the optimal strategy for all agents simultaneously by solving the coupled PDE system given by the HJB~\eqref{eq:HJB} and continuity~\eqref{eq:CE} equations and then use~\eqref{eq:v}. 
Our work follows this macroscopic approach and aims at reducing the immense computational challenges caused by the high dimension of these PDEs.

\subsection*{Potential Mean Field Games}
Assume that there exist functionals $\mathcal{F},~\mathcal{G}:\mathcal{P}(\R^d)\to \R$ such that
\begin{equation*}
F(x,\rho)=\frac{\delta \CF(\rho)}{\delta\rho} (x),\quad G(x,\rho)=\frac{\delta \CG(\rho)}{\delta\rho}(x),
\end{equation*}
where $\frac{\delta}{\delta\rho}$ is the variational derivative; i.e., for a function $w\in L^2(\R^d)$ and probability measure $\rho(x)dx\in \mathcal{P}(\R^d)$ we have that
\begin{equation*}
\begin{split}
\lim\limits_{h\to 0} \frac{\mathcal{F}(\rho+hw)-\mathcal{F}(\rho)}{h}=&\int_{\R^d} F(x,\rho) w(x)dx,
\end{split}
\end{equation*}
and similarly for $\mathcal{G}$.

In the seminal paper \cite{LasryLions2007}, Lasry and Lions observed that, in this case, the MFG system~\eqref{eq:HJB} and~\eqref{eq:CE} coincides with first-order optimality conditions of the infinite-dimensional constrained optimization problem 
\begin{equation}\label{eq:minimization}
\begin{split}
 \inf_{\rho, v}\; & \quad \mathcal{J}_{\rm MFG}(v,\rho) \\
 \text{ s.t.} &\quad \partial_t \rho(x,t) + \nabla \cdot (\rho(x,t)v(x,t))=0, \quad t \in (0,T] \\
               &\quad \rho(x,0) = \rho_0(x), \qquad  x \in \R^d
\end{split}
\end{equation}
whose objective functional reads
\begin{equation} \label{eq:JMFG}
    \begin{split}
\mathcal{J}_{\rm MFG}(v,\rho) = & \int_0^T \int_{\R^d}L\left(x,v(x,t)\right)\rho(x,t) dx dt \\ 
&  + \int_0^T \mathcal{F}(\rho(\cdot,t)) dt + \mathcal{G}(\rho(\cdot,T)).        
    \end{split}
\end{equation}
This important class of MFGs is called potential MFGs. 
Here, the optimal strategies for all agents can be found simultaneously by solving the variational problem~\eqref{eq:minimization}. 
This is reminiscent of potential games, whose Nash equilibria are critical points of a single function that is called a potential.
Remarkably,  the Lagrange multiplier associated with the constraint in~\eqref{eq:minimization} satisfies the HJB equation~\eqref{eq:HJB}.

We use~\eqref{eq:v} to re-parameterize the control in~\eqref{eq:minimization} and solve the infinite-dimensional optimal control problem
\begin{equation}\label{eq:mfg_as_optim_Phi}
 \inf_{\rho,\Phi} \mathcal{J}_{\rm MFG}(-\nabla_p H(x,\nabla \Phi),\rho) + \mathcal{C}_{\rm HJB}(\Phi,\rho) \text{ s.t.~\eqref{eq:CE}}
\end{equation}
where we also added a penalty term for the HJB equation that reads
\begin{equation}\label{eq:HJBpenalty}
    \begin{split}
    \mathcal{C}_{\rm HJB}(\Phi,\rho) = &\; \alpha_1 \int_0^T\int_{R^d} C_1(\Phi,\rho,x,t) \rho(x,t) dx dt \\
    & + \alpha_2 \int_{R^d} C_2(\Phi,\rho,x)  \rho(x,T) dx.
    \end{split}
\end{equation}
The terms penalize violations of the HJB equation in $(0,T)$ and at the final time, respectively, and read
\begin{align}
    & C_1(\Phi,\rho,x,t) \nonumber\\
    & =    |\partial_t \Phi(x,t)-H(x,\nabla \Phi(x,t)) + F(x,\rho(x,t))| \label{eq:C1}\\
    & C_2(\Phi,\rho,x) = |\Phi(x,T)-G(x,\rho(x,T))|. \label{eq:C2}
\end{align}
Adding these terms does not affect the minimizer; however, we show in a numerical experiment that it improves the convergence of the discretized problem. 
Since we penalize the violation of the optimality conditions of the original optimization problem, our penalty bears some similarity with Augmented Lagrangian methods.
The square of the first term has also been used in~\cite{YangEtAl2019}.
Our use of the absolute value in both terms is similar to exact penalty methods; see, e.g., ~\cite[Ch. 15]{NocedalWright2006}.
Compared to the equivalent problem~\eqref{eq:minimization}, this formulation enforces~\eqref{eq:v} by construction, which can reduce the computational cost of a numerical solution; see \cite{Chow2019}. 

\subsection*{Mean Field Control}
Mathematically, mean field control (MFC) models are similar to potential mean field games.
The key difference in MFC is that a central planner devises an optimal strategy, $v:\mathbb{R}^d\times [0,T] \to \mathbb{R}^d$.
All agents in the population follow this strategy in~\eqref{eq:z}, resulting in individual cost given by \eqref{eq:single_cost}. 
The goal of the central player to minimize the overall costs obtained by replacing the running cost, $\mathcal{F}$, and final cost, $\mathcal{G}$, in~\eqref{eq:mfg_as_optim_Phi} by
\begin{equation}\label{eq:Ftilde}
    \begin{split}
        \int_{\R^d} F(x,\rho(x))\rho(x)dx   \text{ and }
        \int_{\R^d} G(x,\rho(x))\rho(x)dx,
    \end{split}
\end{equation}
respectively.
This choice changes the right-hand sides in the HJB equation~\eqref{eq:HJB} to 
\begin{align*}
 &    F(x,\rho)+\int_{\R^d} \frac{\delta F}{\delta \rho} (\rho)(x) \rho dx\\
    \text{ and}\quad & 
    G(x,\rho)+\int_{\R^d} \frac{\delta G}{\delta \rho} (\rho)(x) \rho dx.
\end{align*}
These terms are the variational derivatives of the running and final congestion costs in~\eqref{eq:Ftilde}, respectively. 
Similar to potential MFGs, finding the optimal strategy $v$ is equivalent to determining the potential $\Phi$ by solving~\eqref{eq:mfg_as_optim_Phi}.

\section*{Lagrangian Method}
We follow a discretize-then-optimize-approach, to obtain  a finite-dimensional instance of the optimal control problem~\eqref{eq:mfg_as_optim_Phi}.
The key idea of our framework is to overcome the curse of dimensionality by using a Lagrangian method to discretize and eliminate the continuity equation~\eqref{eq:CE} and all other terms of the objective in~\eqref{eq:mfg_as_optim_Phi}.
We note that the trajectories of the individual agents~\eqref{eq:z} are the characteristic curves of the continuity equation~\eqref{eq:CE}.
Hence,  Lagrangian coordinates are connected to the microscopic model and are a natural way to describe MFGs.
Thereby, we obtain a stable and mesh-free discretization that parallelizes trivially.

To keep the discussion concrete and our notation brief, from now on we consider the $L_2$  transport costs
\begin{equation}\label{eq:LL2}
    L(x,v)=\frac{\lambda_{\rm L}}{2} \|v\|^2,
\end{equation}
where $\lambda_{\rm L} >0$ is a fixed parameter. 
Using~\eqref{eq:H}, it is easy to verify that the Hamiltonian is
\begin{equation}\label{eq:Hl2}
    H(x,p)=\frac{1}{2\lambda_{\rm L}}\|p\|^2.
\end{equation}
We emphasize that our framework can be adapted to handle other choices of transport costs.

We solve the continuity equation~\eqref{eq:CE} using the method of characteristics and Jacobi's identity \cite[Ch. 10]{Bellman}.
Thereby we eliminate the density $\rho$ by using the constraint in \eqref{eq:mfg_as_optim_Phi} and obtain an unconstrained problem whose optimization variable is $\Phi$.
Given $\Phi$, we obtain the characteristics by solving the ODE
\begin{equation}\label{eq:characteristic}
    \begin{split}
        \partial_t z(x,t) & = - \nabla_p H (z(x,t),\nabla \Phi(z(x,t),t))\\
                          & = - \frac{1}{\lambda_{\rm L}}\nabla \Phi(z(x,t),t)
    \end{split}
\end{equation}
with $z(x,0)=x$.  
The first step is derived by inserting~\eqref{eq:v} into~\eqref{eq:z} and the second step follows from~\eqref{eq:Hl2}. 
For clarity, we explicitly denote the dependence of the characteristic on the starting point $x$ in the following.
Along the curve $z(x,\cdot)$, the solution of the continuity equation satisfies for all $t \in [0,T]$
\begin{equation}\label{eq:pushFwd}
 \rho(z(x,t),t) \det(\nabla z(x,t)) = \rho_0(x).
\end{equation}
In some cases, e.g., optimal transport, it is known that the characteristic curves do not intersect, which implies that the mapping $x \mapsto z(x,t)$ is a diffeomorphism and the Jacobian determinant is strictly positive~\cite[Lemma 7.2.1]{ambrosio08}.    

Solving the continuity equation using~\eqref{eq:pushFwd} requires an efficient way for computing the Jacobian determinant along the characteristics. 
Direct methods require in general $\CO(d^3)$ floating point operations (FLOPS), which is intractable when $d$ is large.
Alternatively, we follow~\cite{ZhangEtAl2018,YangEtAl2019} and use the Jacobi identity, which characterizes the evolution of the logarithm of the Jacobian determinant along the characteristics, i.e., 
\begin{equation}\label{eq:logDet}
    \begin{split}
        \partial_t l(x,t)     & =  -\nabla \cdot (\nabla_p H (z(x,t),\nabla \Phi(z(x,t),t)))  \\
                             &= - \frac{1}{\lambda_{\rm L}} \Delta \Phi(z(x,t),t)
    \end{split}
\end{equation}
with $l(x,0)=0$. 
Here, the second step uses~\eqref{eq:Hl2}, $\Delta$ is the Laplace operator, and we denote $l(x,t)=\log \left( \det\left(\nabla z (x,t)\right) \right)$.
This way, we avoid computing the determinant at the cost of numerical integration along the characteristics followed by exponentiation; see also~\cite{Bellman}. 
When the determinant is required, we recommend using an accurate numerical integration technique to avoid large errors arising from the exponentiation.
However, we shall see below that many problems can be solved using the log determinant.

An obvious, yet important, observation is that Lagrangian versions of some terms of the optimal control problems do not involve the Jacobian determinant. 
For example, using~\eqref{eq:pushFwd} and applying the change of variable formula to the first term in \eqref{eq:JMFG} gives
\begin{align}
     \int_{\R^d}& L\left(x,v(x,t)\right)\rho(x,t) dx \\
&=  \int_{\R^d}L\left(z(x,t),v(z(x,t),t))\right)\rho_0(x) dx.
\end{align}
Similarly, the Lagrangian versions of the penalty~\eqref{eq:HJBpenalty} do not involve the Jacobian determinant. 

Using this expression, we can obtain the accumulated transport costs associated with $x$ as $c_{\rm L}(x,T) \rho_0(x)$ where $c_{\rm L}$ is given by
\begin{equation}
    \begin{split}
        \partial_t c_{\rm L}(x,t) &  = L(x,-\nabla_p H(x,\nabla \Phi(x,t))) \\
                                 & = \frac{1}{2\lambda_{\rm L}} \| \nabla \Phi(x,t) \|^2.
    \end{split}
\end{equation}
Here, $c_{\rm L}(x,0)=0$ and as before the second step uses~\eqref{eq:Hl2}.

We also accumulate the running costs associated with a fixed $x$ along the characteristics by integrating
\begin{equation*}
    \partial_t c_{\rm F}(x,t) = \hat{F}(z(x,t),\rho(z(x,t),t),t),
\end{equation*}
where $c_{\rm F}(x,\rho_0(x),0)=0$ and $\hat{F}$ denotes the integrand obtained by applying a change of variables to the functional. 
As this computation is application-dependent, we postpone it until the numerical experiment section.

We note that the trajectories associated with an optimal potential $\Phi$ must satisfy the HJB equation~\eqref{eq:HJB}.
One way to ensure this by construction is to integrate the Hamiltonian system as also proposed in~\cite{ZhangEtAl2018}. 
As doing so complicates the use of the Jacobi identity, we penalize violations of the HJB along the characteristics using
\begin{equation}
    \partial_t c_{1}(x,t) = C_1(\Phi,\rho,z(x,t),t), \quad c_{1}(x,0) = 0,
\end{equation}
where the right hand side is given in~\eqref{eq:C1}.
In summary, we compute the trajectories, log-determinant, transportation costs, running costs, and HJB violations by solving the initial value problem
\begin{equation}\label{eq:characteristics}
    \partial_t\begin{pmatrix}
    z(x,t)\\
    l(x,t)\\
    c_{\rm L}(x,t) \\
    c_{\rm F}(x,t) \\
    c_{1}(x,t) \\
    \end{pmatrix}=\begin{pmatrix}
    -\frac{1}{\lambda_{\rm L}}\nabla \Phi(z(x,t),t)\\
    - \frac{1}{\lambda_{\rm L}} \Delta \Phi(z(x,t),t)\\
    \frac{1}{2\lambda_{\rm L}}\|\nabla \Phi(z(x,t),t) \|^2\\
    \hat{F}(z(x,t),t)\\
    C_1(z(x,t),t)\\
    \end{pmatrix}.
\end{equation}
As discussed above, $z(x,0) = x$ is the origin of the characteristic and all other terms are initialized with zero.
We use the first two components of the ODE to solve the continuity equation and the last three terms to accumulate the running costs along the characteristics.

\subsection*{Optimization Problem}
We now approximate the integrals in~\eqref{eq:mfg_as_optim_Phi} using a quadrature rule.
Together with the Lagrangian method~\eqref{eq:characteristics}, this leads to an unconstrained optimization problem in the potential $\Phi$, which we will parameterize using a neural network in the next section.

Our approach is modular with respect to the choice of the quadrature; however, as we are mostly interested in high-dimensional cases, we use  Monte Carlo integration, i.e., 
\begin{equation} \label{eq:mfg_lagrangian_stoch}
    \begin{split}
\min_{\Phi}  \mathbb{E}_{\rho_0} &  \big( c_{\rm L}(x,T) + c_{\rm F}(x,T)  + \hat{G}(z(x,T))
\\
 & + \alpha_1 c_{1}(x,T)  + \alpha_2 C_2(\Phi,\rho, z(x,T))\big),
    \end{split}
 \end{equation}
where the characteristics are computed using~\eqref{eq:characteristics} and the change of variable used to compute $\hat{G}$ are discussed in the numerical experiment section.
The above problem consists of minimizing the expected loss function, which is given by the sum of the running costs, terminal costs, and HJB penalty.
 
Let $x_1, \ldots, x_N \in \R^d$ be random samples from the probability distribution with density $\mu \in \mathcal{P}(\R^d)$; common choices for $\mu$ are uniform distribution or $\mu=\rho_0$. 
Then, summarizing the computations from this section, we obtain the  optimization problem 
\begin{equation} \label{eq:mfg_lagrangian}
    \begin{split}
\min_{\Phi}  \sum_{k=1}^N & v_k \big( c_{\rm L}(x_k,T) + c_{\rm F}(x_k,T)  + \hat{G}(z(x_k,T))
\\
 & + \alpha_1 c_1(x_k,T)  + \alpha_2 C_2(\Phi,\rho, z(x_k,T))\big),
    \end{split}
 \end{equation}
 where the quadrature weight for the measure $I(g) = \int_{\R^d} g \rho_0(x) dx$ associated with the $k$th sample point $x_k$ is
\begin{equation}\label{eq:quad}
    v_k = \frac{\rho_0(x_k)}{\mu(x_k) N}.
\end{equation}

It is worth re-iterating that we transformed the original optimal control problem \eqref{eq:mfg_as_optim_Phi} in $\Phi$ and $\rho$ to an unconstrained optimization problem in $\Phi$.
For a given $\Phi$, we eliminate the constraints by solving~\eqref{eq:characteristics} independently for each point $x_k$.
In our experiments, we use a fourth-order Runge-Kutta scheme with equidistant time steps.
Since the terms of the cost functions can be computed individually, our scheme is trivially parallel.

Optimization algorithms for solving~\eqref{eq:mfg_lagrangian} can roughly be divided into two classes: stochastic approximation~\cite{Robbins:1951ko} and sample average approximation (SAA)~\cite{NemirovskiEtAl2009}; see also the recent survey~\cite{bottou2016optimization}. 
The further class contains, e.g., variants of the stochastic gradient scheme.
These methods aim at iteratively solving~\eqref{eq:mfg_lagrangian_stoch} using a sequence of steps, each of which uses gradient information computed using a relatively small number of sample points. 
A crucial parameter in these methods is the sequence of step sizes (also known as learning rate) that is typically decaying.
When chosen suitably, the steps reduce the expected value in~\eqref{eq:mfg_lagrangian_stoch}; however, it is not guaranteed that there exists a step size that reduces the approximation obtained for the current sample points.
This prohibits the use of line search or trust region strategies and complicates the application of second-order methods.
By contrast, the idea in SAA methods is to use a larger number of points such that~\eqref{eq:mfg_lagrangian} is a suitable approximation of~\eqref{eq:mfg_lagrangian_stoch}. 
Then, the problem can be treated as a deterministic optimization problem and solved, e.g., using line-search or Trust Region methods.
We have experimented with both types of scheme and found an SAA approach based on quasi-Newton schemes with occasional resampling most effective for solving MFG and MFC problems to high accuracy; see the supplementary information (SI) for an experimental comparison.


\section*{Machine Learning Framework for MFGs}
To obtain a finite-dimensional version of~\eqref{eq:mfg_lagrangian}, we parameterize the potential function using a neural network.
This enables us to penalize violations of the HJB equation~\eqref{eq:HJB} during training and thereby ensure that the trained neural network accurately captures the optimality conditions.
Using a neural network in the Lagrangian method gives us a mesh-free scheme.
In this section, we give a detailed description of our neural network models and the computation of the characteristics. 

\subsection*{Neural Network Models for the Potential}
We now introduce our neural network parameterization of the potential. 
While this is not the only possible parameterization, using neural networks to solve high-dimensional PDE problems has become increasingly common; see, e.g.,~\cite{E2016,Han:2016ku,HanEtAl2017,Sirignano:ik}.
The novelty of our approach is the design of the neural network architecture such that the characteristics in~\eqref{eq:characteristics} can be approximated accurately and efficiently.

Our network models map the input vector $s=(x,t)\in\R^{d+1}$ to the scalar potential $\Phi(s)$. 
In the following,  we denote our model as $\Phi(s,\theta)$, where $\theta$ is a vector of network parameters (also called weights) to be defined below.

The neural network processes the input features using a number of layers, each of which combines basic operations such as affine transformations and element-wise nonlinearities.
While the size of the input and output features is determined by our application, there is flexibility in choosing the size of the hidden layers, which is also called their widths.
Altogether, the number, widths, and specific design of the layers are referred to as the network's architecture.

Our base architecture is defined as follows
\begin{equation}\label{eq:PhiNN}
    \begin{split}
    \Phi(s,\theta) = w^\top N(s, \theta_N) + & \hf s^\top \left(A + A^\top\right) s + c^\top s + b, \\         
     & \theta = (w, \theta_N, {\rm vec}(A), c,b)
    \end{split}
\end{equation}
where for brevity, $\theta$ collects the trainable weights $w\in\R^m, \theta_N \in \R^p, A \in \R^{(d+1)\times (d+1)}, c \in \R^{d+1}, b \in \R$.
The function $N$ can be any neural network with $(d+1)$ input features, $m$ output features, and parameters $\theta_N\in\R^p$.
The last two terms allow us to express quadratic potentials easily.
Our approach is modular with respect to the network architecture. 
However, the design of the neural network's architecture is known to affect its expressibility and the ease of training; see, e.g.,~\cite{Li:2017ta}.  
It is important to note that the use of the neural network renders~\eqref{eq:mfg_lagrangian} in general non-convex.

In this work, our network architecture is a residual network (ResNet)~\cite{He:2016tt}.
For a network with $M$ layers and a given input feature $s\in\R^{d+1}$, we obtain $N(s,\theta_N) = u_M$ as the final step of the forward propagation
\begin{equation}\label{eq:ResNN}
    \begin{split}
    u_0 & = \sigma(K_0 s + b_0) \\ 
    u_1 & = u_0 + h \sigma(K_1 u_0 + b_1)\\
    \vdots & \qquad\qquad\vdots\\  
    u_M &= u_{M-1} + h \sigma(K_M u_{M-1} + b_M),\\  
    \end{split}
\end{equation}
where $\sigma:\R \to\R$ is an element-wise activation function, $h>0$ is a fixed step size, and the network's weights are $K_0 \in\R^{m \times (d+1)}$, $K_1,\ldots, K_M \in \R^{m\times m}$, and $b_0, \ldots, b_M \in \R^m$.
In our experiments we use the activation function 
\begin{equation}
    \sigma(x) =\log(\exp(x) + \exp(-x)),
\end{equation}
which can be seen as a smoothed out absolute value. 
For notational convenience, we vectorize and concatenate all weights in the vector $\theta_N \in \R^p$. 
In theory, the larger the number of layers and the larger their width, the more expressive the resulting network. 
In practice, the full expressiveness may not be realized when the learning problem becomes too difficult and numerical schemes find only suboptimal weights.
In our numerical experiments, we found a relatively shallow networks with as little as $M=1$ layers and widths of $16$ to be very effective; however, this architecture may not be optimal for other learning problems.
The main difference between the forward propagation through a ResNet and a more traditional multilayer perceptron is the addition of the previous feature vector in the computation of $u_1, \ldots, u_M$. 

ResNets have been tremendously successful in a wide range of machine learning tasks and have been found to be trainable even for large depths (i.e., $M\gg0$). 
By interpreting~\eqref{eq:ResNN} as a forward Euler discretization of an initial value problem, the continuous limit of a ResNet is amenable to mathematical analysis~\cite{E:2017kz,HaberRuthotto2017}.
This observation has received a lot of attention recently and been used, e.g., to derive maximum principle~\cite{Li:2017wr},  propose stable ResNet variants~\cite{HaberRuthotto2017}, and to accelerate the training using multi-level~\cite{Chang:2017te} and parallel-in-time schemes~\cite{GuntherEtAl2018}.

\subsection*{Characteristics Computations}
To compute the characteristic and other quantities in~\eqref{eq:characteristics}, we need to compute the gradient and Laplacian of $\Phi$ with respect to the input features. 
The perhaps simplest option is to use automatic differentiation (AD), which has become a ubiquitous and mature technology in most machine learning packages. 
We note that this convenience comes at the cost of $d$ separate evaluations of directional derivatives when computing the Laplacian.
While trace estimation techniques can lower the number of evaluations, this introduces inaccuracies in the PDE solver. 
Also, we show below that computing the Laplacian exactly is feasible for our network.

We now provide a detailed derivation of our gradient and Laplacian computation. 
The gradient of our model in~\eqref{eq:PhiNN} with respect to the input feature $s$ is given by
\begin{equation}\label{eq:DPhi}
    \nabla_s \Phi(s,\theta)  = \nabla_s N(s,\theta_N) w + (A + A^\top) s + c.
\end{equation}
Due to the ordering in $s=(x,t)$, the first $d$ component of this gradient correspond to the spatial derivatives of the potential, and the final one is the time derivative.
We compute the gradient of the neural network~\eqref{eq:ResNN} in the direction $w$ using back-propagation (also called reverse mode differentiation)
\begin{equation}\label{eq:backprop}
\begin{split}
    z_M     &= w + h K_M^\top {\rm diag}(\sigma'(K_M u_{M-1} + b_M)) w,\\  
    \vdots & \qquad \qquad\vdots\\
    z_{1} &= z_2 + h K_{1}^\top {\rm diag}(\sigma'(K_{1} u_{0} + b_{1})) z_2,\\  
    z_{0} &= K_{0}^\top {\rm diag}(\sigma'(K_{0} s + b_{0})) z_1,\\  
\end{split}
\end{equation}
which gives $\nabla_s N(s,\theta_N) w = z_0$.  Here, ${\rm diag}(v) \in\R^{m\times m}$ is a diagonal matrix with diagonal elements given by $v\in\R^m$ and $\sigma'(\cdot)$ is computed element-wise. 

Next, we compute the Laplacian of the potential model with respect to $x$.
We first note that
\begin{equation*}
    \Delta \Phi(s,\theta) =  {\rm tr}\left(E^\top ( \nabla_s^2 ( N(s,\theta_N) w) + (A+A^\top) )E \right),
\end{equation*}
where the columns of $E \in \R^{(d+1)\times d}$ are given by the first $d$ standard basis vectors in $\R^{d+1}$.
Computing the terms involving $A$ is trivial, and we now discuss how to compute the Laplacian of the neural network in one forward pass through the layers.
We first note that the trace of the first layer in~\eqref{eq:ResNN} is
\begin{equation}
\begin{split}
    t_0  & = {\rm tr}\left(E^\top  \nabla_s ( K_0^\top {\rm diag}(\sigma'(K_0 s + b_0)) z_1)  E\right)\\
         & = {\rm tr}\left(E^\top K_0^\top {\rm diag}(\sigma''(K_0 s + b_0) \odot z_1) K_0 E \right)\\
         & = (\sigma''(K_0 s + b_0) \odot z_1)^\top ((K_0 E)\odot(K_0 E)) \mathbf{1},
\end{split}
\end{equation}
where $\odot$ denotes a Hadamard product (i.e., an element-wise product of equally-sized vectors or matrices), $\mathbf{1}\in\R^{d}$ is a vector of all ones,  and in the last we used that the middle term is a  diagonal matrix.
The above computation shows that the Laplacian of the first layer can be computed using $\CO(m\cdot d)$ FLOPS by first squaring the elements in the first $d$ columns of $K_0$, then summing those columns, and finally one inner product. 
The computational costs are essentially equal to performing one single matrix-vector product with the Hessian using AD but produces the exact Laplacian.

To compute the Laplacian of the entire ResNet, we continue with the remaining rows in ~\eqref{eq:backprop} in reverse order to obtain
\begin{equation}\label{eq:LapResNet}
    \Delta ( N(s,\theta_N) w) = t_0 + h \sum_{i=1}^M t_i,
\end{equation}
where $t_i$ is computed as
 \begin{equation*}
    \begin{split}
        t_i & = {\rm tr}\left(J_{i-1}^\top \nabla_s ( K_i^\top {\rm diag}(\sigma'(K_i u_{i-1}(s) + b_i)) z_{i+1})J_{i-1}\right) \\
           & = {\rm tr}\left(J_{i-1}^\top K_i^\top {\rm diag}(\sigma''(K_i u_{i-1} + b_i) \odot z_{i+1}) K_i J_{i-1} \right)\\
           & = (\sigma''(K_i u_{i-1} + b_i) \odot z_{i+1})^\top ((K_i J_{i-1})\odot(K_i J_{i-1})) \mathbf{1}.
    \end{split}
\end{equation*}
Here, $J_{i-1} = \nabla_s u_{i-1}^\top \in \R^{m\times d}$ is a Jacobian matrix, which can be updated and over-written in the  forward pass at a computational cost of $\CO(m^2 \cdot d)$ FLOPS. 

To summarize the above derivations, we note that each time step of the numerical approximation of the characteristics involves one forward propagation~\eqref{eq:ResNN}, one back-propagation~\eqref{eq:backprop} and the trace computations~\eqref{eq:LapResNet}. 
The overall computational costs scale as $\CO(m^2 \cdot d \cdot M)$, i.e., linearly with respect to the input dimension and number of layers, but quadratically with the width of the network. 
This motivates the use of deep and not wide architectures.

\section*{Numerical Experiments}
We apply our method to two prototype MFG instances.
Here, we give a concise overview of the problems and results and refer to SI~\ref{SI:num} for a more detailed description of the problem instances and results.
We perform our experiments  using a prototype of our algorithm that we implemented in the Julia programming language~\cite{Bezanson:2017gd} as an extension of the machine learning framework Flux~\cite{innes:2018}.
We publish the code under a permissible open source license at \url{http://github.com/EmoryMLIP/MFGnet.jl}.

\subsection*{Example 1: Dynamical Optimal Transport}
Given two densities $\rho_0, \rho_1 \in \CP(\R^d)$, the Optimal Transport (OT) problem consists of finding the transformation $y: \R^d \to \R^d$ with the smallest transport costs such that the push forward of $\rho_0$  equals $\rho_1$.
The problem was introduced by Monge \cite{BenamouBrenier2000}, revolutionized by the contributions by Kantorovich~\cite{Kantorovich2006},  Benamou and Brenier~\cite{BenamouBrenier2000}. Other notable theoretical advances include~\cite{Evans1997,Ambrosio2003,Villani2003,villani2008optimal}.  

Among the many versions of OT, we consider the fluid dynamics formulation~\cite{BenamouBrenier2000}, which can be seen as a potential MFG.
This formulation is equivalent to a convex optimization problem, which can be solved accurately and efficiently if $d\leq3$; see, e.g.,~\cite{HaberHoresh2015}.
The curse of dimensionality limits applications of these methods when $d>3$. 
Approximately solving high-dimensional OT problems is of key interest to machine learning and Bayesian statistics; see some related works in~\cite{YangEtAl2019}.

We model the fluid dynamic version of optimal transport in~\cite{BenamouBrenier2000} as a potential mean field game by using
\begin{equation*}
  \CF(\rho)=0,  \quad \CG(\rho) = \lambda_{\rm KL} \mathcal{G}_{\rm KL}(\rho),
\end{equation*}
where $\lambda_{\rm KL}>0$ is a penalty parameter, and  the second term is the Kullback Leibler divergence, which penalizes violations of the final time constraint $\rho(\cdot,T)=\rho_1(\cdot)$ and reads
\begin{equation}\label{eq:GKL}
    \begin{split}
    \mathcal{G}_{\rm KL}(\rho) &=  \int_{\R^d} \rho(x,T)\log \frac{\rho(x,T)}{\rho_1(x)} dx 
    \\
    =  \int_{\R^d} & \log \frac{\rho(z(x,T),T)}{\rho_1(z(x,T))} \rho_0(x)dx    
    \\
    =  \int_{\R^d}&\left( \log\rho_0(x)-l(x,T)-\log \rho_1(z(x,T))\right) \rho_0(x)dx.
    \end{split}
\end{equation}
Here, the log determinant, $l$, is given in~\eqref{eq:logDet}.
The variational derivative of this loss function, which is used in the computation of the penalty term in~\eqref{eq:C2}, is
\begin{equation*}
	G_{\rm KL}(x,\rho) = 1 + \log(\rho(x)) -\log \rho_1(x).
\end{equation*}
Note that the optimal control problem~\eqref{eq:mfg_as_optim_Phi} is symmetric with respect to the roles of $\rho_0$ and $\rho_1$ if and only if $\lambda_{\rm KL}=\infty$ because we relaxed the final time constraint.

\begin{figure}
    \begin{center}
        \includegraphics[width=0.45\textwidth]{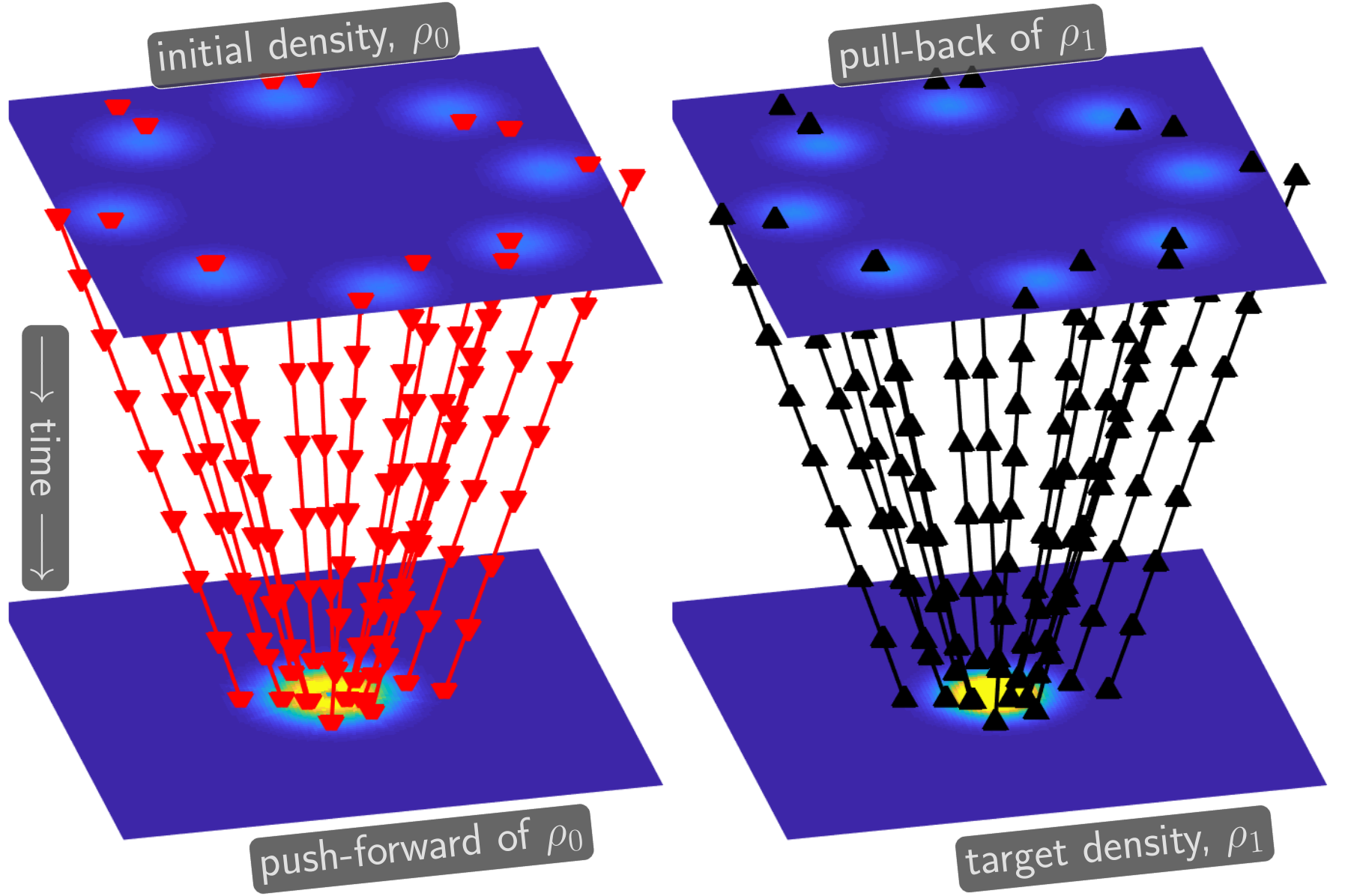}
    \end{center}
    \caption{Visualization of a two-dimensional optimal transport experiment. The left column shows the given initial density $\rho_0$ (top) and its push-forward at time $t=1$ (bottom). The red lines represent the characteristics starting from randomly sampled points according to $\rho_0$. The right column shows the given target density, $\rho_1$, (bottom) and its pull-back (top). The black line corresponds to the characteristics computed backward in time from the endpoints of the red characteristics. The similarity of the images in each row and the fact that the characteristics are almost straight lines show the success of training. Also, since the red and black characteristics are nearly identical, the transformation is invertible. The same color axis is used for all plots.}
    \label{Fig2}
\end{figure}

We generate a synthetic test problem that enables us to increase the dimension of the problem with only minimal effect on its solution.
For a given dimension $d$, we choose the density of the Gaussian white noise distribution as the target
\begin{equation*}
    \rho_1(x) = \rho_G\left(x,\mathbf{0}, 0.3\cdot\bfI\right).
\end{equation*}
Here, $\rho_G(\cdot,\bfm,\bfSigma)$ is the probability density function of a $d$-variate Gaussian with  mean $\bfm\in\R^d$ and covariance matrix $\bfSigma\in\R^{d\times d}$. 
The initial density is the Gaussian mixture     
\begin{equation*}
    \rho_0(x) = \frac{1}{8} \sum_{j=1}^{8} \rho_G \left(x,\bfm_j, 0.3 \cdot \bfI\right),
\end{equation*}
where the means of the individual terms are equally spaced on the intersection of the sphere with radius four and the coordinate plane in the first two space dimensions, i.e., 
\begin{equation*}
    \bfm_j = 4 \cdot \cos\left(\frac{2\pi}{8} j \right) \bfe_1 + 4 \cdot \sin\left(\frac{2\pi}{8} j \right)\bfe_2, \quad j=1,\ldots,8.
\end{equation*}
Here, $\bfe_1$ and $\bfe_2$ are the first two standard basis vectors.
The two-dimensional instances of these densities are visualized in Fig.~\ref{Fig2}.

We perform four experiments to show the scalability to high-dimensional instances, explore the benefits of the penalty function $\mathcal{C}_{\rm HJB}$, compare two optimization strategies, and validate our scheme by comparing it to a Eulerian method in $d=2$, respectively.
The network architecture is almost exactly the same in all cases; see the supplementary information (SI) for details. 

We show in Fig.~\ref{Fig2} that our network provides a meaningful solution for the two-dimensional instance.
The performance can be seen by the close match of the given $\rho_0$ and the pull-back of $\rho_1$, which is optimized as it enters the terminal costs, and the similarity of the push-forward of $\rho_0$ and $\rho_1$, which is computed after training. Also, the characteristics are approximately straight.
To improve the visualization, we compute the characteristics using four times as many time integration steps as used in training.
A close inspection of the push-forward density also shows that the network learns to partition the target density into eight approximately equally-sized slices, as predicted by the theory; see Fig.~\ref{fig:highDimPlots} in the SI.

Our method obtains quantitatively similar results in higher-dimensional instances ($d=10,50,100$) as can be seen in the summary of the objective function values provided in the top half of Tab.~\ref{tab1}.
The values are computed using the validation sets. 
The table also shows that despite a moderate growth of the number of training samples, the actual runtime per iteration of our prototype implementation per iteration grows slower than expected from our complexity analysis.
We used more samples in higher dimensions to avoid over-fitting; see Fig.~\ref{fig:highDimConv} in the SI.
Due to the design of the problem, similar transport costs and terminal costs are to be expected. 
There is a slight increase of the terminal costs for the $d=50$-dimensional instances, but we note that, at least for projections onto the first two coordinate dimensions, the image quality is similar for all cases; see Fig.~\ref{fig:highDimPlots} in the SI.

In our second experiment, we show that without the use of the HJB penalty, $\mathcal{C}_{\rm HJB}$, the optimization can fail to match the densities and result in curved characteristics, which are not meaningful in OT; see Fig.~\ref{figHJB} in the SI.
Increasing the number of time steps to discretize the characteristics~\eqref{eq:characteristics}, improves the results, however, the results are still inferior to the ones obtained with the penalty, and the computational cost of training is four times higher.
Hence, in this example, using the penalty function, we obtain a more accurate solution at reduced computational costs. 

Our third OT experiment compares two optimization strategies: the SAA approach with BFGS used throughout our experiments and an SA approach with ADAM that is common in machine learning. 
We note that, for the two-dimensional instance of our problem, ADAM converges considerably slower and is less effective in reducing the HJB penalty at a comparable computational cost; however,  both methods lead to similar final results.

In our fourth experiment, we compare our proposed method to a provably convergent Eulerian solver for the $d=2$-dimensional instance. 
To this end, the optimal controls obtained using both methods are compared using an explicit Finite-Volume solver for the continuity equation, which was not used during the optimization.
This step is essential to obtain a fair comparison.
In Fig.~\ref{fig:EulerianOT}, Fig.~\ref{fig:EulerianOTPhi},  and Tab.~\ref{tab:Eulerian} we show that, for this example, our method is competitive and, as demonstrated above, scales to  dimensions that are beyond reach with Eulerian schemes.

\begin{table}[t]
    \begin{center}
\begin{tabular}{@{}|@{\;}c@{\;}|@{\;}c@{\;}|@{\;}c@{\;}|@{\;}c@{\;}|@{\;}c@{\;}|@{\;}c@{\;}|@{\;}c@{\;}|@{}} 
            \hline
			\multicolumn{7}{|c|}{Example 1: Optimal Transport} \\ \hline
            $d$ & $N$   &  $\mathcal{L}$  & $\mathcal{F}$& $\mathcal{G}$ & $\mathcal{C}_{\rm HJB}$ & time/iter (s)\\ \hline
            2  & 2,304  &  9.99e+00  & - &   7.01e-01   &   1.17e+00  &  2.038    \\
            10 & 6,400  &  1.01e+01  & - &   8.08e-01   &   1.21e+00  &  8.256    \\
            50 & 16,384 &  1.01e+01  & - &   6.98e-01   &   2.94e+00  &  81.764   \\
            100& 36,864 &  1.01e+01  & - &   8.08e-01   &   1.21e+00  &  301.043  \\ \hline
			\multicolumn{7}{|c|}{Example 2: Crowd Motion} \\ \hline
            $d$ & $N$   &  $\mathcal{L}$ & $\mathcal{F}$& $\mathcal{G}$ & $\mathcal{C}_{\rm HJB}$ & time/iter (s)\\ \hline
            2  & 2,304  &  1.65e+01      & 2.29e+00 &   7.81e-01  &   2.27e-01  &  4.122    \\
            10 & 6,400  &  1.65e+01      & 2.22e+00 &   7.51e-01  &   3.94e-01  &  17.205   \\
            50 & 9,216  &  1.65e+01      & 1.91e+00 &   7.20e-01  &   1.21e+00  &  134.938  \\
            100& 12,544 &  1.65e+01      & 1.49e+00 &   1.00e+00  &   2.78e+00  &  241.727  \\   \hline     
	 \end{tabular}
    \end{center}
    \caption{Overview of numerical results for instances of the optimal transport and crowd motion problem in growing space dimensions. All values were approximated using the validation points.  {\normalfont Legend: $d$ (space dimension), $N$ (number of training samples), $\mathcal{L}$ (transport costs), $\mathcal{F}$ (running costs), $\mathcal{G} $ (terminal costs), $\mathcal{C}_{\rm HJB}$ HJB penalty } }
    \label{tab1}
\end{table}

\subsection*{Example 2: Crowd Motion}
We consider the motion of a crowd of agents distributed according to an initial density $\rho_0$ to the desired state $\rho_1$. 
In contrast to the OT problem, the agents trade off reaching $\rho_1$ with additional terms that encode their spatially varying preference and their desire to avoid crowded regions. 

To force the agents to move to the target distribution, we use the same terminal cost as in the previous example.  
To express the agents' preference to avoid congestion and model costs associated with traveling through the center of the domain, we use the mean field potential energy
\begin{equation}\label{eq:MFGF}
\mathcal{F}(\rho(\cdot,t))=  \lambda_{\rm E} \CF_{\rm E}(\rho(\cdot,t)) + \lambda_{\rm P} \CF_{\rm P}(\rho(\cdot,t)),
\end{equation}
which is a weighted sum of an entropy and preference term defined, respectively, as  
\begin{align*}
    \CF_{\rm E}(\rho(\cdot,t)) & = \int_{\R^d} \rho(x,t)\log\rho(x,t) dx, \\
    \CF_{\rm P}(\rho(\cdot,t)) & = \int_{\R^d} Q(x)\rho(x,t)dx.
\end{align*}
The entropy terms penalizes the accumulation of agents; i.e., the more spread out the agents are, the smaller this term becomes.
In the third term, $Q:\R^d \to\R$, models the spatial preferences of agents; i.e., the smaller $Q(x)$, the more desired is the position $x$. 
Carrying out similar steps to compute these terms in Lagrangian coordinates, we obtain
\begin{align}
    \CF_{\rm E}(\rho(\cdot,t)) & = \int_{\R^d} (\log \rho_0(x) - l(x,t)) \rho_0(x) dx, \\
    \CF_{\rm P}(\rho(\cdot,t)) & = \int_{\R^d} Q(z(x,t))\rho_0(x)dx. 
\end{align}
In our experiment, we control the relative influence of both terms by choosing $\lambda_{\rm E} = 0.01,  \lambda_{\rm P}=1$ in~\eqref{eq:MFGF}, respectively.
To penalize the $L_2$ transport costs, we use the Lagrangian and Hamiltonian given in ~\eqref{eq:LL2} and~\eqref{eq:Hl2}.

Finally, we take the variational derivative to obtain the HJB equation~\eqref{eq:HJB}. We note that the mean field coupling is
\begin{equation*}
    \frac{\delta \CF(\rho)}{\delta \rho}    
     = F(x,\rho) =  \lambda_{\rm F} F_{\rm E}(x,\rho) + \lambda_{\rm P} F_{\rm P}(x,\rho),
\end{equation*}
with the terms
\begin{align*}
    F_{\rm E}(x,\rho) & = \log \rho(x) +1,\\
    F_{\rm P}(x,\rho) & = Q(x).
\end{align*}

\begin{figure}[t]
    \begin{center}
        \includegraphics[width=.35\textwidth]{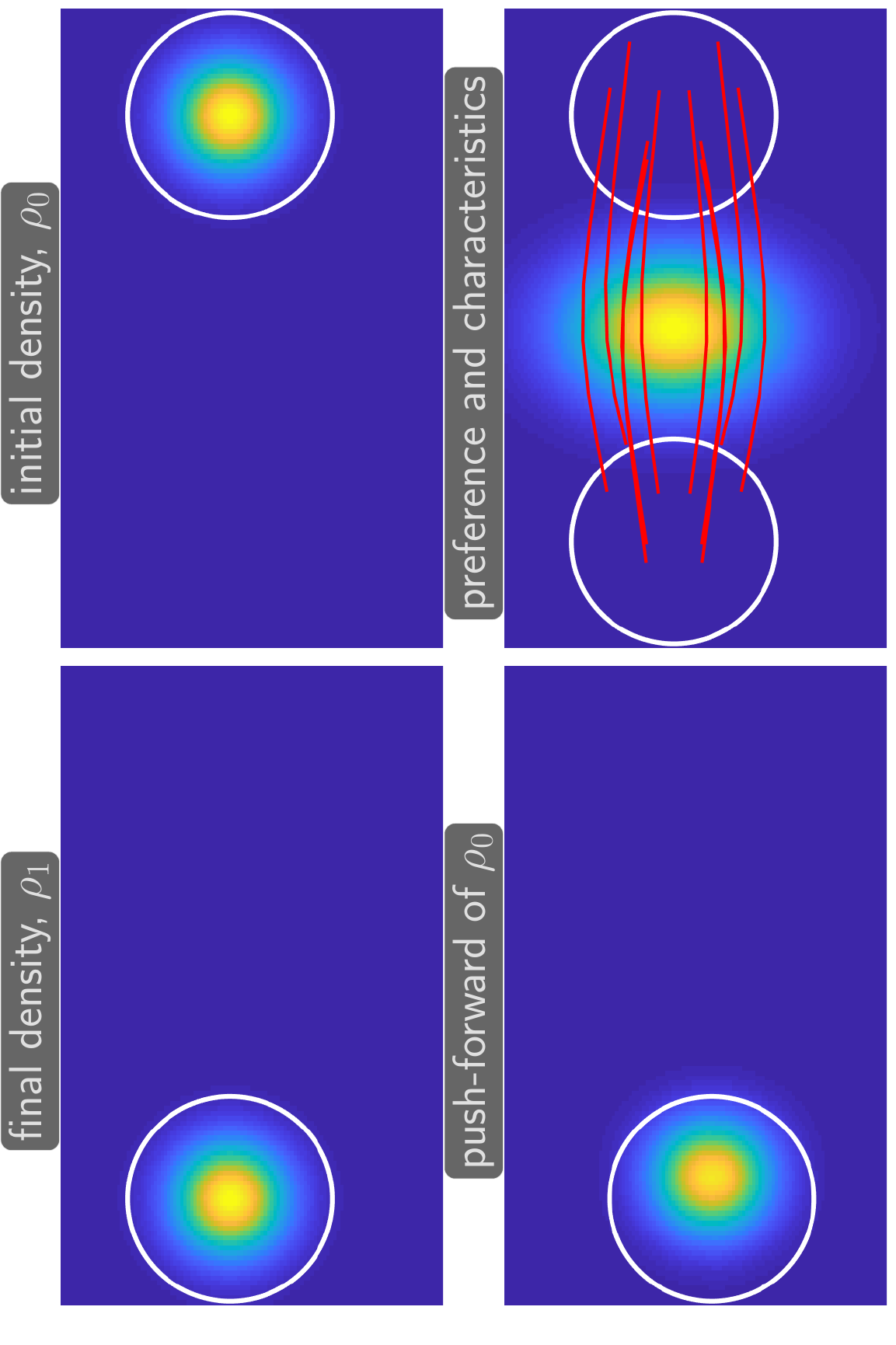}
    \end{center}
    \caption{Illustration of the two-dimensional crowd motion problem. The initial density (top left) and target density (bottom left) are Gaussians centered at the top or bottom of the domain, respectively. The white circles depict a contour line for the density and are added in the remaining subplots to help visual assessment. The agents' goal is to move from their initial distribution approximately to the target while avoiding congestion and the obstacle in the center of the domain (top right). There are higher costs associated with traveling through regions where the preference function is large (represented by yellow regions).  The red lines in the right subplot show the learned trajectories. It can be seen that the characteristics are curved to avoid the center of the domain. We also show that the push-forward of the initial density is similar to the target density (bottom right). }
    \label{fig:MFG}
\end{figure}

A detailed description of the experimental setup is provided in the SI.
The initial and target densities are shifted Gaussians 
\begin{equation*}
    \rho_0 = \rho_G(x,3 \cdot \bfe_2, 0.3 \cdot \bfI), \quad \rho_1(x) = \rho_G(x, -3 \cdot \bfe_2, 0.3\cdot \bfI).
\end{equation*}
Note that for $\lambda_{\rm E} = \lambda_{\rm P} = 0$, the problem becomes a trivial OT problem and the optimal trajectories would be straight and parallel through the center of the domain.
To obtain a more interesting dynamic, we introduce the preference function
\begin{equation*}
    Q(x) = 50 \cdot \rho_G\left(x,\mathbf{0}, {\rm diag}(1, 0.5)\right).
\end{equation*}
Since $Q$ attains its maximum at the origin, agents have an incentive to avoid this region, which leads to curved characteristics.
As terminal costs, we use the Kullback-Leibler divergence~\eqref{eq:GKL}.
For the higher-dimensional instances, we evaluate the preference function using the first two components of $x$.
Thereby, the preference function becomes invariant to the remaining entries and the solutions become comparable across dimensions.

As in the OT example, we obtain similar but not identical objective function values for the problem instances obtained by choosing $d \in \{2,10,50,100\}$; see bottom half of Tab.~\ref{tab1}.
Again, we increased the number of training samples with dimension, and we observe that the runtime of our prototype code scales better than predicted.
Fig~\ref{fig:MFG} shows the optimized trajectories for the $d=2$ instance; see Fig~\ref{fig:MFGdim} in the SI for similar visualizations for the remaining instances.
As expected, the agents avoid the crowded regions and prefer to spread out horizontally to avoid congestion.
We chose the starting points of the characteristics to be symmetric about the $x_2$-axis.
As expected, this renders  the learned trajectories approximately symmetric as well.
We note that the characteristics are visually similar across dimensions and that our results are comparable to those obtained with a provably convergent Eulerian solver for the $d=2$-dimensional instance; see Fig.~\ref{fig:EulerianMFG}, Fig.~\ref{fig:EulerianMFGPhi}, and Tab.~\ref{tab:Eulerian} in the SI.

\section*{Discussion and Outlook}
By combining Lagrangian PDE solvers with neural networks, we develop a new framework for the numerical solution of potential mean field games and mean field control problems.
Our method is geared toward high-dimensional instances of these problems that are beyond reach with existing solution methods.
Since our method is mesh-free and well-suited for parallel computation, we believe it provides a promising new direction toward much-anticipated large-scale applications of MFGs. 
Even though our scheme is competitive to Eulerian schemes based on convex optimization for the two-dimensional examples its main advantage is the scalability to higher dimensions.
We exemplify the effectiveness of our method using an optimal transport and a crowd motion problem in 100 dimensions.
Using the latter example, we also demonstrate that our scheme can learn complex dynamics even with a relatively simple neural network model.

The fact that our framework transforms mean field games into novel types of machine learning problems brings exciting opportunities to advance MFG application and theory. 
While we can already leverage the many advances made in machine learning over the last decades, the learning problem in MFGs has some unique traits that require further attention.

Open mathematical issues include the development of practical convergence results for NNs in optimal control. 
In fact, it is not always clear that a larger network will perform better in practice. 
Our numerical experiment for the optimal transport problem makes us optimistic that our framework may be able to solve HJB equations in high dimensions to practically relevant accuracy.
Similarly, the impact of regularization parameters on the optimization deserves further attention.
However, more analysis is needed to obtain firm theoretical results for this property.

A critical computational issue is a more thorough parallel implementation, which may provide the speed-up needed to solve more realistic MFG problems.
Also, advances in numerical optimization for deep learning problems may help solve the learning problems more efficiently.

Open questions on the machine learning side include the set-up of the learning problem. 
There are little to no theoretical guidelines that help design network architectures that generalize well.
Although there is a rich body of examples for data science applications, our learning problem has different requirements.
For example, not only does the neural network need to be evaluated, but the Lagrangian scheme also requires its gradient and Laplacian. 
A careful choice of the architecture may be necessary to ensure the required differentiability of the network. 
Also, more  experiments are needed to establish deeper intuition into the interplay between the network architecture and other hyper-parameters, e.g., the choice of the penalty parameter or the time discretization of the characteristics. 

An obvious open question from an application perspective is the use of our framework to solve more realistic problems.
To promote progress in this area, we provide our code under a permissible open source license.

%% file: acknowledgements.tex
We would like to thank Derek Onken for fruitful discussions and his help with the Julia implementation and  Wonjun Lee and Siting Liu for their assistance with the implementation of the Eulerian solver for the two-dimensional problem instances and their suggestions that helped improve the manuscript. 
L.R. is supported by the National Science Foundation (NSF) Grant No. DMS-1751636.
This research was performed while L.R. was visiting the Institute for Pure and Applied Mathematics (IPAM), which is supported by the NSF Grant No. DMS-1440415.
 S.O., W.L., L.N., S.W.F. receive support from AFOSR MURI FA9550-18-1-0502, AFOSR Grant No. FA9550-18-1-0167, and ONR Grant No. N00014-18-1-2527.

%% file: appendix.tex
\begin{appendix}
\section{Supplemental Information for Numerical Experiments} 
\label{SI:num}
We implemented a prototype version of our framework as an extension of the machine learning framework Flux~\cite{innes:2018} that is written in Julia~\cite{Bezanson:2017gd}. 
Rather than giving a full description of the code here, we publish the code  at
\begin{center}
    http://github.com/EmoryMLIP/MFGnet.jl
\end{center}
We provide unit tests to ensure the validity of the code.

We performed our experiments on a Microway workstation running Ubuntu Linux.
The system has four Intel Xeon E5-4627 CPUs with a total of 40 cores and 1 TB of memory. 
The system is shared with other users, which is why we report runtimes as averages over the entire training as a rough measure of the computational cost. 
Our code is not optimized for runtime, and additional speedups can be achieved by exploiting the parallelism provided by our method and by using accelerated hardware such as Graphics Processing Units (GPU).

\subsection*{Dynamical Optimal Transport}
\label{app:OT}

\paragraph{Experimental Setup. } 
We use the ResNet model described in~\eqref{eq:ResNN} with a width of $m=16$ and $M=1$ time step with a step size of $h=1$.
To be precise, this means that the ResNet weights are $K_0 \in \R^{16 \times (d+1)}$, $K_1 \in \R^{16\times 16}$, and $b_0, b_1 \in \R^{16}$. 
Note that only the width of the first layer changes with the dimension of the problem instance.
The elements in the matrices $K_0$ and $K_1$ are initialized by sampling from $\mathcal{N}(0,0.01)$ and the components of the vectors $b_0$ and $b_1$ are drawn from $\mathcal{N}(0,0.1)$. 
The terms of the quadratic parts in our model~\eqref{eq:PhiNN}, $A, c, b$ are initialized with zeros, and the vector $w\in\R^{16}$ is initialized with a vector of all ones.

In the mean field objective function, the transport costs are multiplied with $\lambda_{\rm L} = 2$ and the terminal costs are multiplied with $\lambda_{\rm KL} = 5$.  
The penalty parameters for $\mathcal{C}_{\rm HJB}$ are $\alpha_1=3$ and $\alpha_2=3$. 

Unless noted otherwise, we train our networks using an SAA approach and re-sample the training set after every 25 iterations to reduce the risk of over-fitting. 
Since the number of optimization variables is relatively small (the number of weights is between 637 and 12,223), we use BFGS with a backtracked Armijo line search.
Our implementation follows the presentation in~\cite{NocedalWright2006}.  
To avoid an indefinite Hessian approximation that may arise with this simple line search, we skip the update when the curvature along the current direction is negative. 
Note that in the SAA framework, other optimization methods can be used easily. 
We use a simple multilevel strategy in which we apply 500 iterations using a relatively small training set and then another 500 iterations using the number of training samples reported in Tab.~\ref{tab1}.

We compute the characteristics using a fourth-order Runge-Kutta scheme.
Our default is to use only $n_t = 2$ time steps with constant time step size of $1/2$. 
Using only two time steps is motivated by saving computational time and the known property of the optimal solution to have straight characteristics, which are easy to integrate accurately.
We note, though, that this property only holds for the optimal solution, which is to be found using the neural network training.
Indeed, we show below that using two few time steps may cause the model to learn curved characteristics and that this risk can, in our example, be avoided by using the HJB penalty.

\begin{figure*}[t]
    \begin{center}
        \includegraphics[width=.9\textwidth]{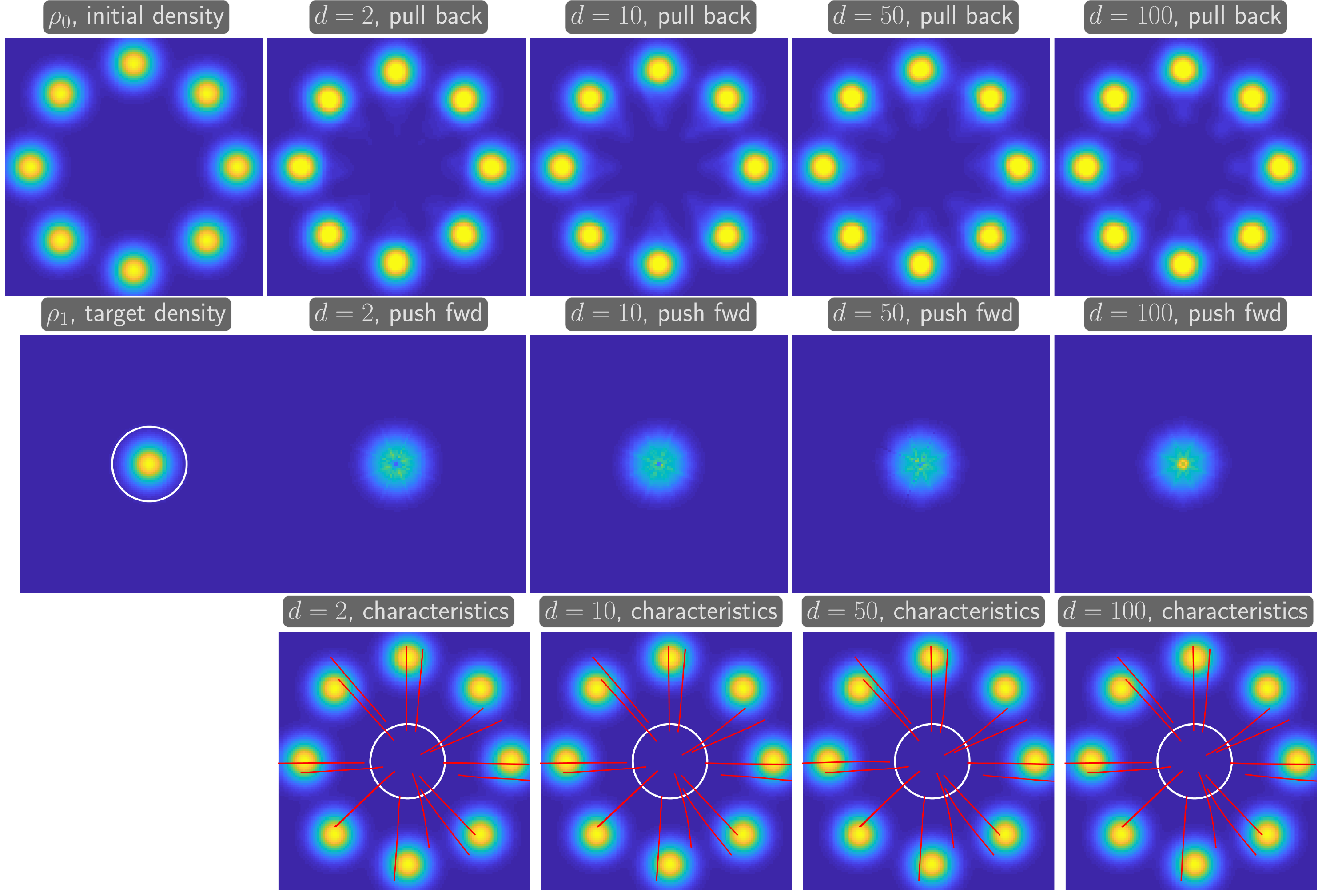}
    \end{center}
    \caption{Visualization of the training results for the optimal transport test problem in dimensions $2, 10, 50, 100$ (left to right). For $d>2$, we show slices along the first two coordinate directions. The left column shows the initial (first row) and target density (second row) on which we superimpose contour lines for reference. The remaining images in the first row show the pull-back of the target density computed with the trained network.  As expected, since the distance between the pull-back density and the initial density is part of the objective function, the images are visually almost identical. While the difference between the push-forward of the initial density and the target (shown in the second row) is slightly larger, we note that the distance is not computed during the optimization. The last row shows projections of the characteristics from 16 randomly chosen points from the initial density. We observe that those points move toward the target (see white contour line) along an almost straight trajectory.  }
    \label{fig:highDimPlots}
\end{figure*}

Our experiments focus on the scalability to higher dimensions and in-depth comparisons of numerical schemes. 
Therefore, we generate a test problem whose solution is similar across different dimensions and use it throughout our experiments.
Two-dimensional projections of the initial and target densities can be seen in the left column of Fig.~\ref{fig:highDimPlots}.
The initial density is the Gaussian mixture obtained by averaging eight $d$-variate Gaussians whose means are equally spaced on the intersection of the sphere with radius four and the $x_1-x_2$ plane. 
The final density is a Gaussian centered at the origin.
The covariance matrix of all Gaussians in the initial and final densities is $0.3 \cdot \bfI$. 
Due to the design of the initial and final density, we expect similar solutions and objective values.


\paragraph{Experiment 1: Scalability.} 
We solve $2, 10, 50, 100$ dimensional instances of our OT test problem using the BFGS approach described above.
At each iteration, we monitor the sample average approximation (SAA)~\eqref{eq:mfg_lagrangian} of the expected value in~~\eqref{eq:mfg_lagrangian_stoch} using a validation set consisting of $4,096$ samples drawn independently from the initial density.
We choose the number of validation samples to be $1024$ for the 2-dimensional case, and $4,096$ for the $10,50,$ and $100$ dimensional cases, respectively.
The training points used during the approximation are drawn independently from the initial density. 
The quadrature weights are computed using~\eqref{eq:quad}.
It is known that the performance of SAA approaches depends on how well the sample average  approximates the expected value.
As expected, we observed that using too few training points leads to overfitting, which can be seen by a large discrepancy between the approximation of the objective function computed using the training and validation points.
Therefore, we increase the number of points in the training data as the dimension grows; see Tab.~\ref{tab1} for details.

\begin{figure*}[t]
    \begin{center}
        \includegraphics[width=\textwidth]{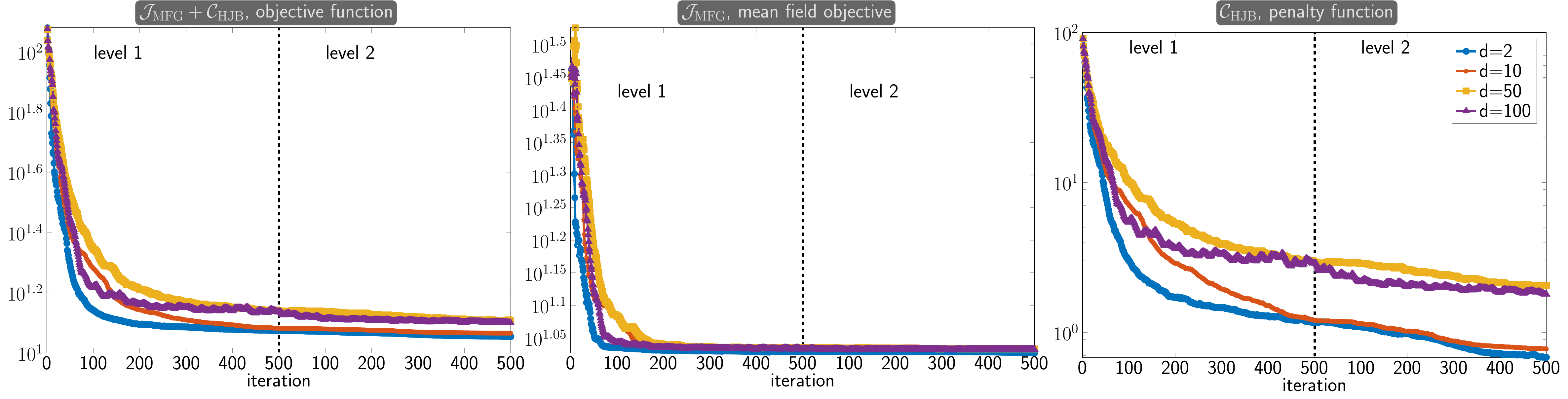}
    \end{center}
    \caption{Comparison of the convergence of the training algorithm for the optimal transport test problem in dimensions $2, 10, 50, 100$. The plot on the left shows the value of the objective function evaluated at each iteration using the validation points. All convergence plots are subdivided into the first 500 iterations performed using fewer training samples and the last 500 iterations with a more accurate approximation.} It can be seen that the relative reduction of the objective function is larger for the smaller dimensional instances. The remaining plots show convergence of the two terms of the objective that are associated with the MFG and the HJB penalty, respectively. These plots show that the difference in the convergence mostly stems from the HJB penalty, which decreases more slowly for the high-dimensional instances in this example.
    \label{fig:highDimConv}
\end{figure*}

We visualize the push-forward of the initial density, the pull-back of the final density, and the characteristics for the different dimensions in the columns of Fig.~\ref{fig:highDimPlots}. 
Also, we provide convergence plots for the objective function, the mean field term, and the HJB penalty in Fig.~\ref{fig:highDimConv}.
For $d>2$, we show image slices along the first two coordinate directions and project the characteristics (red lines) onto the plane given by the first two coordinate dimensions.
As expected, since we compute the terminal costs using the distance between the pull-back density and the initial density, the images are visually almost identical; see also reduction of $\mathcal{J}_{\rm MFG}$ in Fig.~\ref{fig:highDimConv}. 
The difference between the push-forward of the initial density, which is not optimized directly, and the target (shown in the second row) is slightly larger. 
The last row shows projections of the characteristics starting in 16 points randomly chosen from the initial density. 
We observe that those points move toward the target (indicated by a white contour line) along an almost straight trajectory.  
Comparing the different columns of Fig.~\ref{fig:highDimPlots}, it is noticeable that qualitatively similar results are obtained for all spatial dimensions, which indicated the scalability of our method in this case.
A close inspection of the push-forward densities also indicates that the network learns
to partition the target density into eight approximately equally sized slices.

\begin{figure*}[t]
    \begin{center}
        \begin{tabular}{c}
        \includegraphics[width=.7\textwidth]{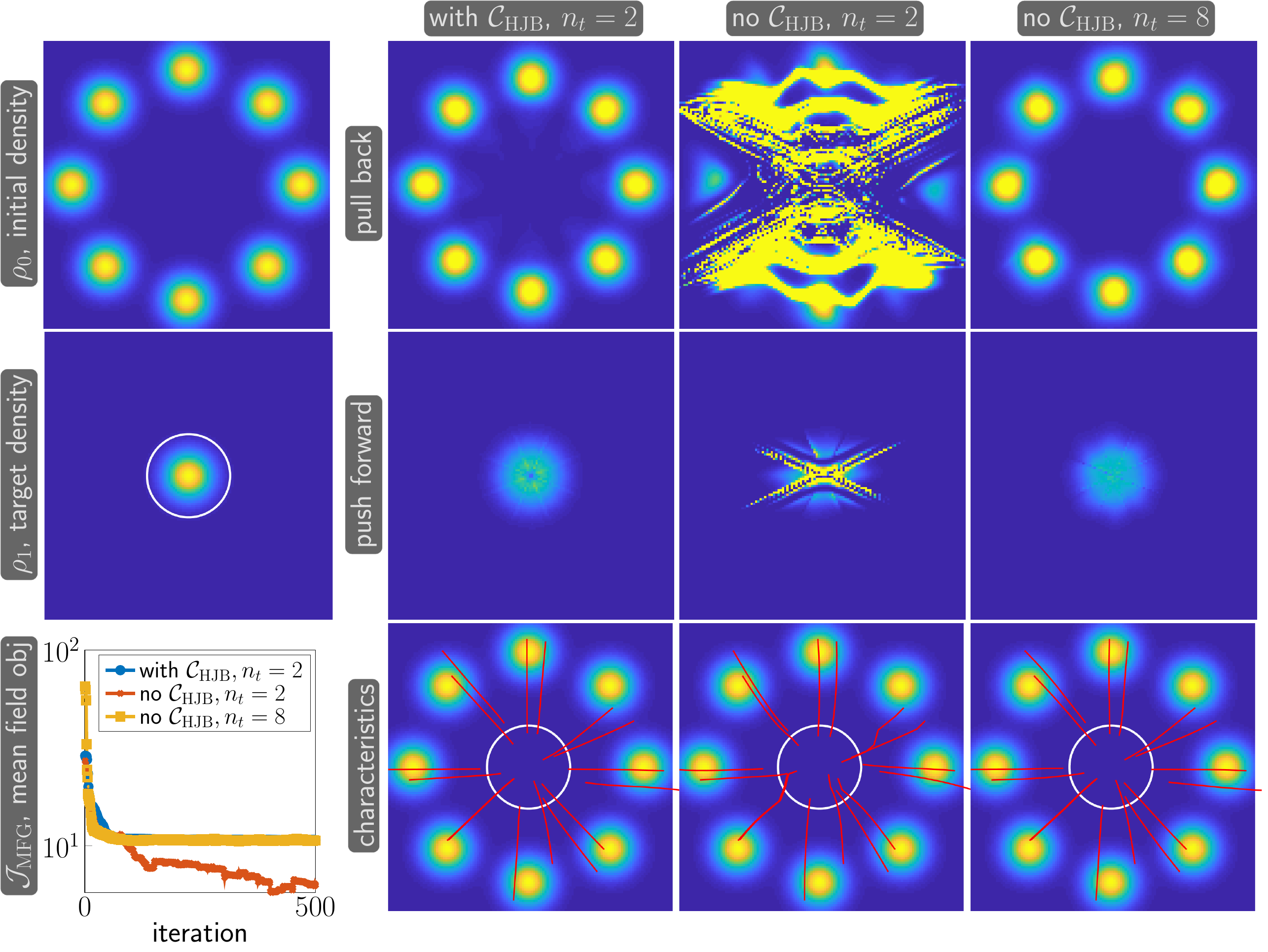}\\
        \end{tabular}
    \end{center}
    \caption{Examining the impact of the HJB penalty for the $d=2$ dimensional instance of the optimal transport problem. The left column shows the initial (first row) and target densities (second row, superimposed by contour lines for orientation) as well as a convergence plot (third row) that shows the mean field objective function approximated using the validation set at the first 500 iterations. The second column visualizes the results obtained with the HJB penalty and $n_t=2$ time steps for the Runge-Kutta time integrator. It can be seen that the pull-back of the final density is almost identical to $\rho_0$ (first row),  the push-forward of $\rho_0$ matches $\rho_1$ (second row), and the characteristics are almost straight (third row). The third column visualizes the results obtained by repeating the previous experiment without the HJB penalty. Although the reduction of the mean field objective is larger (bottom left plot), neither the pull-back nor push-forward densities appear similar to their respective targets. Also, the characteristics are not straight (third row). In the fourth row it can be seen that adding more time steps to the time integrator (in this case $n_t=8$) provides improved but, visually inferior, results to our proposed scheme, which is about four times less expensive computationally.}
    \label{figHJB}
\end{figure*}

\paragraph{Experiment 2: Benefits of $\mathcal{C}_{\rm HJB}$.} 
To assess the benefits of the HJB penalty, $\mathcal{C}_{\rm HJB}$, we compare the results of the two-dimensional instance of the OT problem from the previous experiment to two experiments that do not involve the penalty $\mathcal{C}_{\rm HJB}$.
We show the convergence of the mean field term $\mathcal{J}_{\rm MFG}$ and visualizations of the push-forward and pull-back densities in Fig.~\ref{figHJB}.

In the first experiment, we use the identical experimental setting but deactivate the penalty. 
As can be seen in the lower-left subplot in Fig.~\ref{figHJB} (red line), the optimization reduces the mean field objective substantially. 
However, the plots in the third column show that the network does not solve the problem very well. 
Neither the pull-back nor push-forward densities appear similar to their respective target, and the characteristics (computed here with eight time steps) are not straight. 

In the second experiment, we increase the number of time steps for computing the characteristics to $n_t=8$. 
This quadruples the computational cost of the training but provides more meaningful results; see the fourth column in Fig.~\ref{figHJB}. 
We attribute this also to the increased accuracy of the approximate transport costs in~\eqref{eq:characteristics}, which leads to higher costs when the characteristics are not straight.
Since $\mathcal{C}_{\rm HJB}$ penalizes the optimality conditions of the original problem, this experiment suggests that the penalty becomes less important when using a better time discretization.
These results are similar the ones obtained with the HJB penalty and $n_t=2$ time steps, which is computationally more efficient.

\paragraph{Experiment 3: BFGS vs. ADAM.} 
We use the $d=2$ dimensional instance of the OT problem to investigate the choice of the training method and solve the learning problem in Experiment 1 with the stochastic approximation scheme ADAM~\cite{kingma2014adam}.
ADAM is a commonly used method in machine learning and has also been used for high-dimensional PDE problems in~\cite{HanEtAl2017,Sirignano:ik}.

\begin{figure*}[t]
    \begin{center}
        \includegraphics[width=\textwidth]{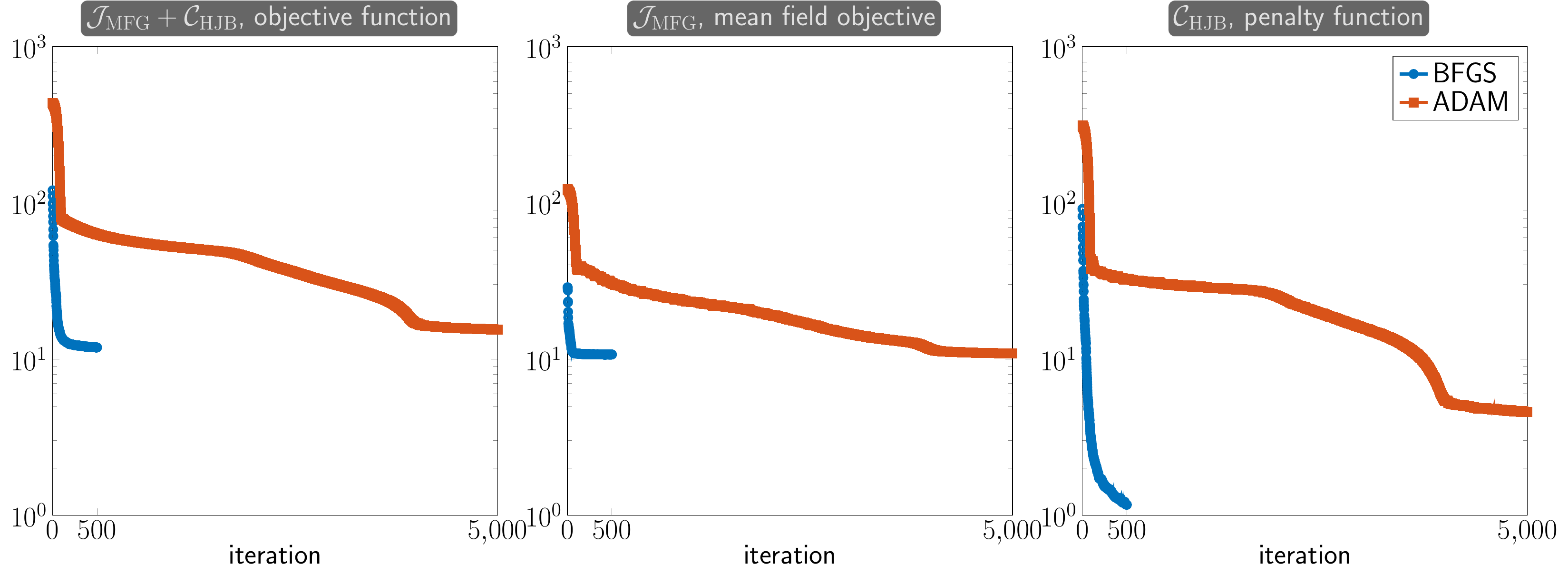}
    \end{center}
    \caption{Convergence of BFGS and ADAM methods for the $d=2$ dimensional instance of the optimal transport problem. We show the convergence of the overall objective (left) and its two terms associated with the mean field game cost (center) and the HJB penalty (right) computed using the validation set. It can be seen that the BFGS scheme converges in fewer iterations, and in particular leads to a more drastic reduction of the HJB penalty.}
    \label{fig:BFGSvsADAM}
\end{figure*}

For ADAM, we perform 5,000 iterations with a batch size of $1,024$ training points sampled from $\rho_0$ and another 5,000 iterations with 2,304 samples. 
We experimented with the batch size and the number of steps performed with a batch before re-sampling.
In our tests, re-sampling the batch every 25 steps was the most effective strategy in terms of validation accuracy. 
The traditional way of drawing a new batch after every step of the method provided slightly worse performance.
As before, we use 1,024 validation points to monitor the objective function, the mean field term, and the HJB penalty at each iteration. 
The convergence plots in Fig.~\ref{fig:BFGSvsADAM} show that for the first 500 iterations in this case, BFGS converges in fewer iterations and, particularly, reduces the HJB penalty more drastically.
Even when the same number of training points is used, the cost per iteration is higher for BFGS than for ADAM due to the Hessian approximation and line searches.
However, the overall costs should be approximately comparable.

\subsection*{Crowd Motion} 
\label{sub:crowd_motion}

We use the same ResNet model, and in the initialization strategy as in the OT experiment, i.e., the width is $m=16$, we use $M=1$ time step with a step size of $h=1$.
Also, we use the same training method, i.e., an SAA version using BFGS, re-discretizing the objective function every 25 steps. 
As before, we solve $2, 10, 50, 100$ dimensional instances of the problem that are designed to lead to a comparable solution.

A notable difference is that we use more time steps in the Runge-Kutta scheme that we use to approximate the characteristics~\eqref{eq:characteristics}. 
Here, we use $n_t=4$ steps with time step size of $1/4$, since we expect the characteristics to bend around the obstacle. 

As terminal costs, we use the Kullback-Leibler divergence~\eqref{eq:GKL} with a weight of $\lambda_{\rm KL} = 5$. 
The parameters for the HJB penalty are $\alpha_1 = 10 $ and $\alpha_2 = 1$. 

\begin{figure*}[t]
    \begin{center}
        \includegraphics[width=.78\textwidth]{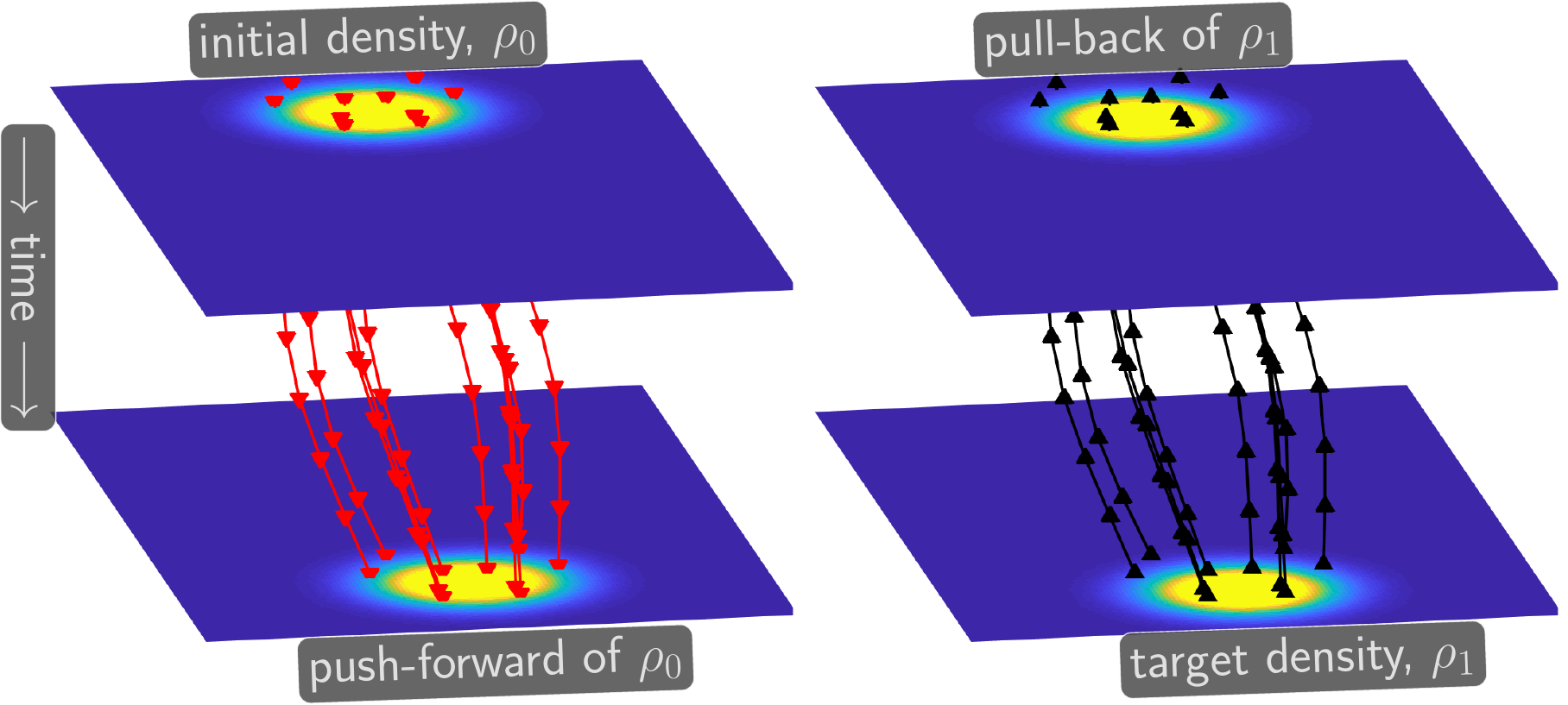}
    \end{center}
        \caption{Visualization of the two-dimensional crowd motion problem. The initial density (top left) and target density (bottom right) are shifted Gaussians with identical covariance.
     The push-forward of $\rho_0$ at time $t=1$ is shown in the bottom left. 
     The red lines represent the characteristics starting from points randomly sampled from $\rho_0$. 
     The pull-back of $\rho_1$ is shown in the top right. 
     The black lines correspond to the characteristics computed backward in time from the end points of the red characteristics. 
     The similarity of the images in each row and the fact that the characteristics are curved to avoid the center of the domain indicate the success of training. Also, since the red and black characteristics are nearly identical, the transformation is invertible. The same color axis is used for all plots. }
	 \label{fig:MFGextra}
\end{figure*}

An alternative visualization to the $d=2$-dimensional results in Fig.~\ref{fig:MFG} in the main text is provided in Fig.~\ref{fig:MFGextra}. 
Here it is noticeable that the characteristics bend to avoid congestion and the center of the domain. 
These two terms are modeled by the running costs $\mathcal{F}$ and are the main difference between the OT problem; for the latter, the characteristics would be straight and parallel.

\begin{figure*}
	\begin{center}
		\includegraphics[width=0.97\textwidth]{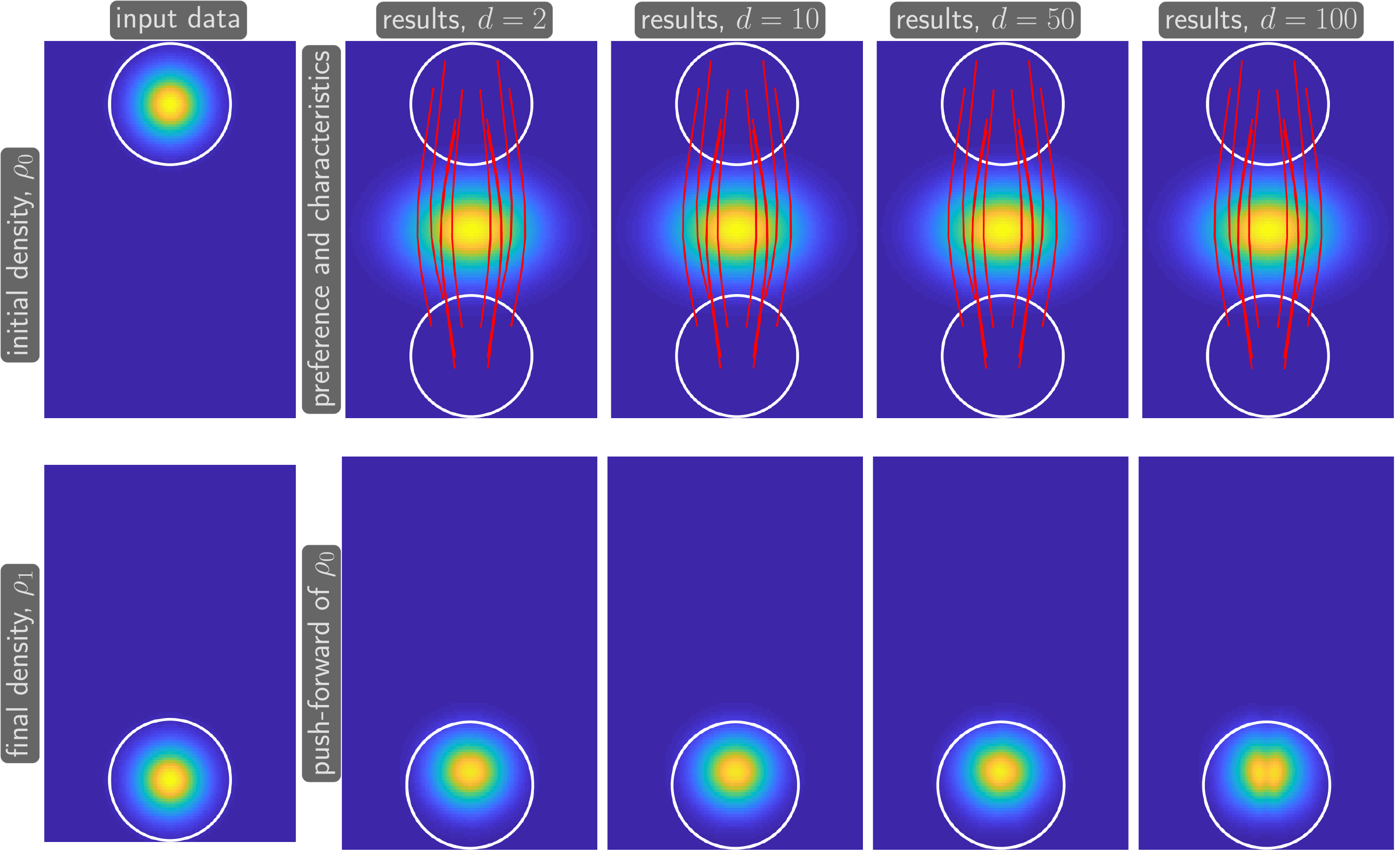}		
	\end{center}
	\caption{Comparison of the $d=2,10,50,100$ dimensional instances of the crowd motion problem (column-wise). The agents' goal is to move from the initial density (top left plot) approximately to the target density (bottom left plot). The white circles depict contour lines of the given densities for better comparison. The remaining columns in the top row show the preference function (which is invariant to $d$) and the characteristics, whose start points are sampled symmetrically about the $x_2$ axis. We use the same origin for the characteristics in each example. The characteristics indicate that, for all cases, the agents learn to avoid the center of the domain, where bright yellow colors indicate high costs. Also, the characteristics are approximately symmetric. The plots in the bottom row show the push-forward of the initial density, $\rho_0$, which in all cases looks similar to the final density.    }
	\label{fig:MFGdim}
\end{figure*}

In Fig.~\ref{fig:MFGdim}, we compare the results across the tested dimensions. 
Comparing the corresponding plots row-wise shows that, as expected by our choice of example, our model learns similar dynamics.
Since we placed an obstacle in the center of the domain (bright yellow colors in the top row are associated with large travel costs) the agents avoid this region.
For visualization, we sampled five starting points of the characteristics from $\rho_0$ and then mirrored them about the $x_2$ axis.
To make the plots comparable, we use the same starting points for all examples, which means we pad the vectors with zeros when $d>2$.
This shows that the learned characteristics are approximately symmetric. 
The bottom row shows the similarity the push-forwards of $\rho_0$ to the target density.
Note that a perfect match is not expected, since agents trade-off transport and running costs with the terminal costs.

\subsection*{Comparison to Eulerian Methods} 
\label{sec:compfv}
We numerically compare our machine learning framework to a state-of-the-art Eulerian method using the $d=2$ dimensional instance of the dynamical optimal transport and the crowd motion problem. 
We obtain the Eulerian scheme by adapting the fluid dynamics approach for the dynamical OT problem in~\cite{HaberHoresh2015} to model the variational formulation of the mean field game in~\eqref{eq:minimization}. 
To this end, we remove the final time constraint $\rho(x,1) = \rho_1(x)$, implement the Kullback-Leibler term in~\eqref{eq:GKL} to quantify the terminal costs, and add the running costs in~\eqref{eq:JMFG}.
Optimizing the momentum instead of the velocity, as originally proposed in~\cite{BenamouBrenier2000}, we obtain a convex optimization problem consisting of a smooth objective function and linear equality constraints that model the dynamics.
Since it is based on a convex formulation of the problem, we consider the solution obtained using the Eulerian scheme as a gold standard. 
Note that there is, to the best of our knowledge, no analytic solution for our problem instances.

We use the discretize-then-optimize approach proposed in~\cite{HaberHoresh2015} to approximately solve the variational problem.
To be precise, we use a staggered discretization of the momentum and density on a regular grid of the space-time domain. 
An advantage of this construction is its stability and that the resulting scheme is conservative; i.e., the numerical solution satisfies the mass-preservation constraint $\partial_t \int_{\Omega} \rho(\cdot,t) dx$ for all $t$.
After discretization, we obtain a finite-dimensional convex optimization problem with linear equality constraints, to which we apply a primal-dual Newton scheme that uses a backtracked Armijo linesearch on the primal and dual residuals.
The backtracking is added to ensure sufficient descent in the violation of the optimality condition but also the positivity of the density.  
As a stopping criterion we require that the Euclidean norm of the primal residual is less than $10^{-8}$ and the norm of the dual residual is less than $10^{-2}$.

We use a coarse-to-fine hierarchy of grids to reduce the computational effort. 
The key idea of this multilevel scheme is to limit the number of fine mesh iterations by computing an starting guess obtained by refining results from a coarser level.
This is particularly attractive in Newton-type methods that benefit from good initialization and is  common practice, e.g., in image processing~\cite{Modersitzki2009}.
We start the optimization on a coarse mesh consisting of only $16\times 16\times 8$ where the first two dimensions are the spatial dimension and the third corresponds to time. 
To obtain a starting guess for the momentum and density, we apply 10 Conjugate Gradient iterations to the linear system given by the equality constraints and threshold this regularized solution to ensure that the density is strictly positive.
The Lagrange multipliers are initialized with zeroes.
In our experiments, the primal-dual Newton scheme provides an accurate approximation of the global minimizer on this discretization level.
We then re-discretize the problem on the next finer grid, which consists of $32\times32\times 16$ cells, and solve the optimization problem starting with the interpolated solution from the coarse grid. 
We repeat this procedure until we reach the fine grid of $128\times 128\times 64$, where the problem has about 3 million unknowns.

We implement the above finite volume scheme as an extension of the MATLAB toolbox FAIR~\cite{Modersitzki2009} and provide the codes in the Github repository associated with this paper.

For the optimal transport problem, the coarse-grid iteration terminates after 25 Newton steps at an iterate whose norm of the primal and dual residuals are reduced from $7.8\cdot10^{-2}$ and $1.9\cdot10^{4}$ to $5.5\cdot 10^{-17}$ and $1.6\cdot 10^{-3}$, respectively.
The number of iterations required to achieve similar reductions on the intermediate levels are 13 and 15.
On the finest level, we perform only 10 iterations and reduce the primal and dual residuals by about 14 and 5 orders of magnitude, respectively.
For the crowd motion problem,  the number of iterations required on the respective grids are 144, 23, 17, and 5.
From these final fine-mesh iterates, we then compute the optimal controls, i.e., the velocities on a staggered space-time grid with $128\times 128\times 64$ cells.

To provide a direct comparison between our Lagrangian machine learning framework and the Eulerian scheme, we solve the continuity equation with the optimal controls on a fine grid and numerically approximate the values of the objective functional.
It is important to solve the problems using the exact same numerical scheme, since different discretization will in general provide different approximations of the objective function.
Here, we use an explicit finite volume scheme for solving the continuity equation.
We obtain the optimal control of the machine learning framework by evaluating the trained neural network $\Phi(x,t)$ for points $(x,t)$ chosen on the same space-time grid used in the Eulerian scheme.
Based on these estimates, we use tri-linear interpolation to solve the continuity equation on a mesh with $256\times 256$ pixels in space and $512$ time steps.
The number of time steps is chosen small enough to satisfy the CFL condition.
As before, we use a conservative finite volume scheme that, in exact arithmetics, preserves the mass of the density.

\begin{figure}
	\begin{center}
		\includegraphics[width=0.5\textwidth]{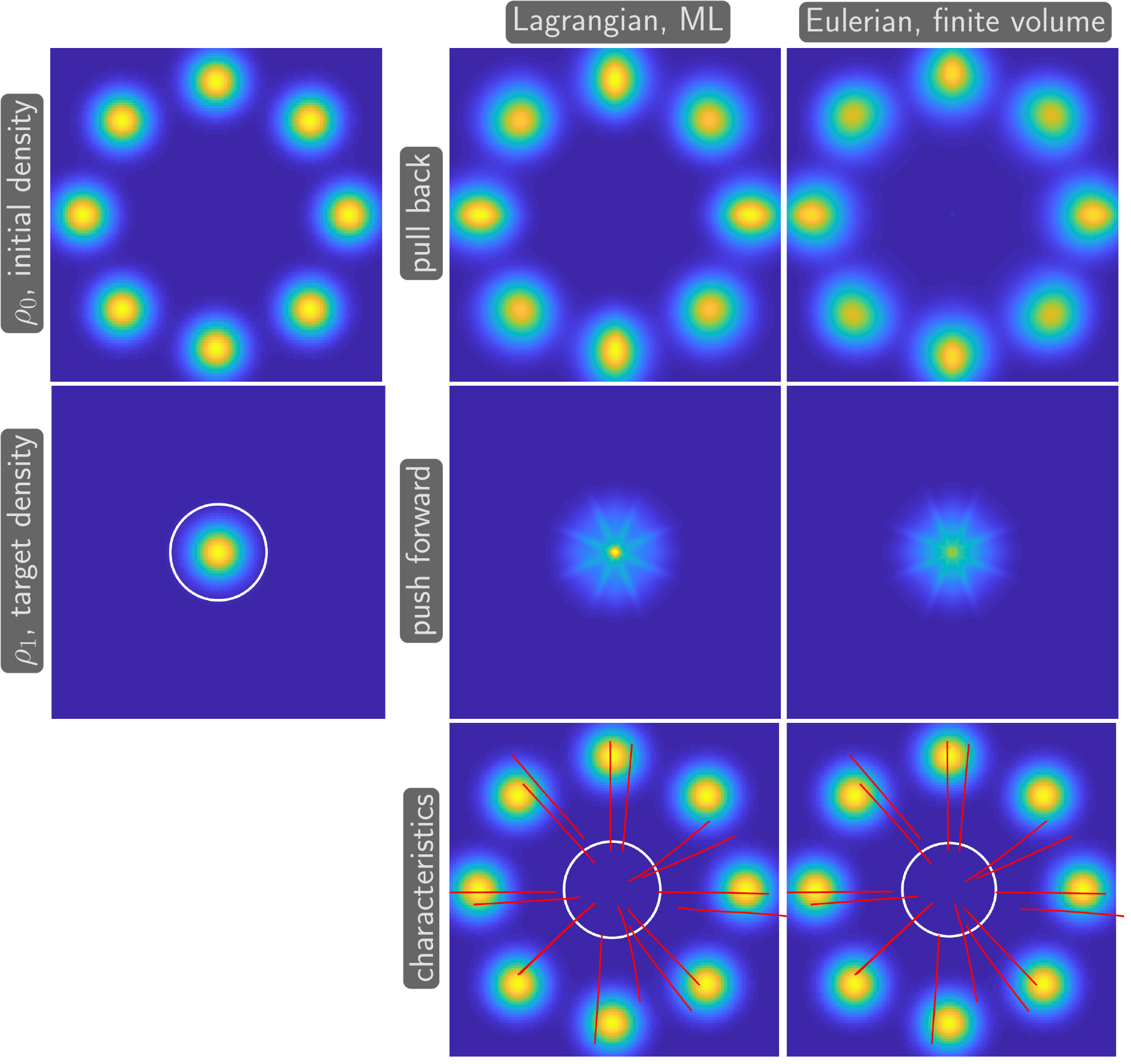}
	\end{center}
	\caption{Comparison of the $d=2$ dimensional optimal transport problem computed using the Lagrangian machine learning framework and a provably-convergent Eulerian technique. 
    The left column shows the initial (first row) and target density (second row, superimposed by a white contour line for orientation).
    The center column shows the pull-back, push-forward, and the characteristics computed using the velocity provided by the proposed Lagrangian machine learning framework with $n_t=2$. 
    We compute the densities by solving the continuity equation on a grid using explicit time-stepping and conservative finite volume discretization. 
    We apply the same scheme to the optimal velocities determined by the Eulerian scheme, which is based on convex programming and thus provably convergent; see right column.
    Both schemes provide visually comparable results, and the overall objective function is about $0.59$\% higher for the machine learning framework. 
    Upon close inspection one can see that the image similarity is higher for the ML scheme, but the transport costs are lower for the Eulerian; see also Tab.~\ref{tab:Eulerian}
	}
	\label{fig:EulerianOT}
\end{figure}

\begin{figure}
	\begin{center}
		\includegraphics[width=0.4\textwidth]{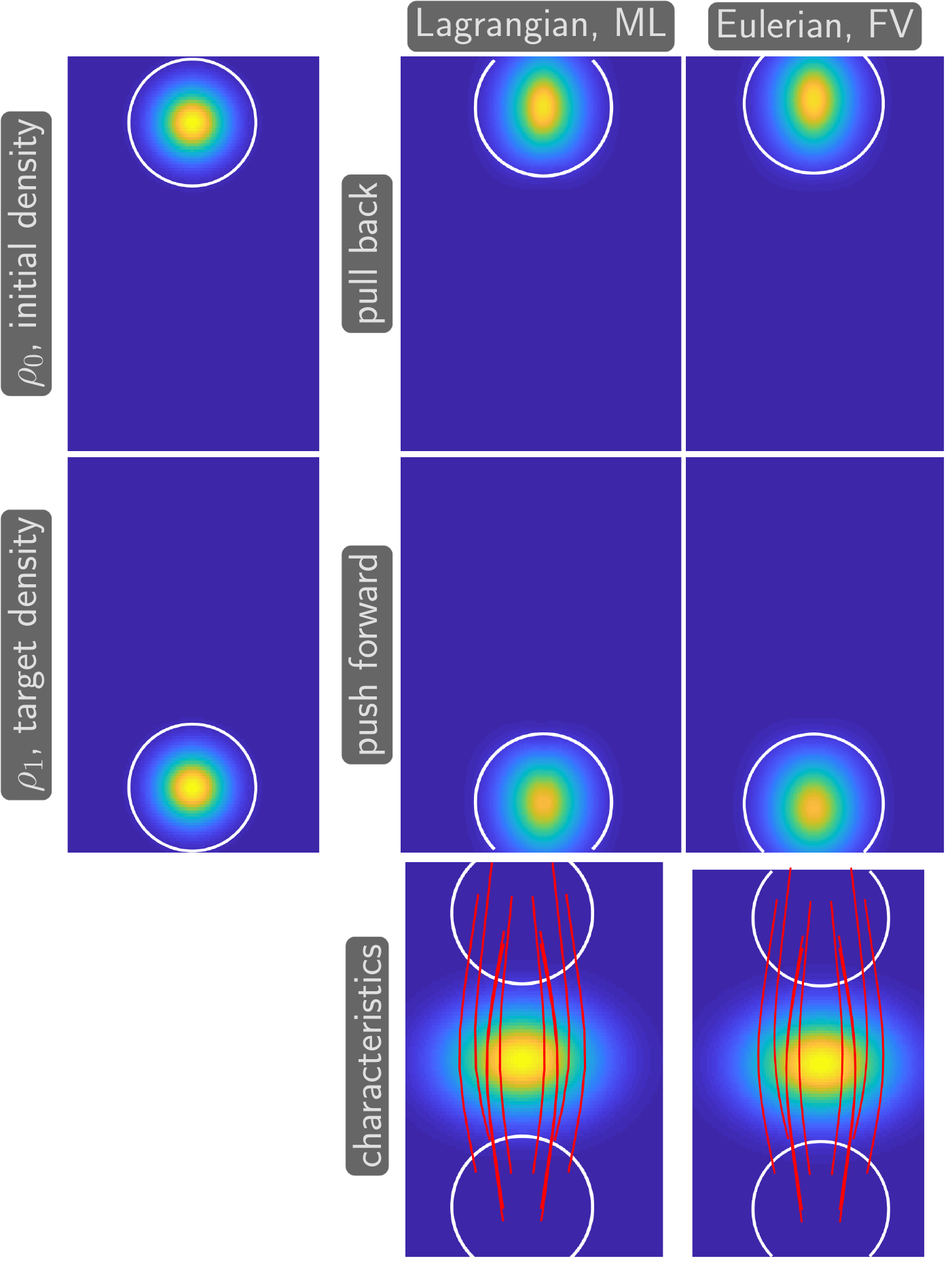}
	\end{center}
	\caption{Comparison of the optimal control computed using our Lagrangian machine learning framework and an Eulerian technique based on the dynamic formulation of the $d=2$-dimensional crowd motion problem. 
    The left column shows the initial (first row) and target density (second row), each superimposed by a white contour line for orientation.
    The center column shows the pull-back (top), push-forward (center), and the characteristics (bottom) computed using the velocity provided by the proposed Lagrangian machine learning framework. 
    The plot of the characteristics also shows the preference function, which assigns higher costs for travel through the center of the domain.    
    We obtain the densities by solving the continuity equation on a grid using explicit time-stepping and conservative finite volume discretization. 
    We use the same PDE solver with the optimal velocities computed by the Eulerian scheme; see right column.
    Both schemes provide visually comparable results, and the overall objective function is about $0.39$\% higher for the machine learning framework; see also Tab.~\ref{tab:Eulerian}.
	}
	\label{fig:EulerianMFG}
\end{figure}

In Tab.~\ref{tab:Eulerian}, we show the transport, running, and terminal costs approximated by the finite-volume scheme for the optimal transport and crowd motion problem.
For the Eulerian schemes we report the results obtained at the two finest levels and we include all settings of the Lagrangian method listed above.
Overall, we see that the various instances of the Lagrangian OT methods are highly competitive (and in some examples superior) to the fine mesh OT solution.
Surprisingly, the overall lowest loss value is observed for the Lagrangian scheme with $n_t=8$ time steps and no HJB penalty; however, including $\mathcal{C}_{\rm HJB}$ and using only $n_t=2$ time steps is less than 1\% sub-optimal and superior to the results obtained with the Eulerian scheme at half resolution. 
This is a remarkable result due to the non-convexity of the training and the vastly reduced number of parameters ($\approx$ 3 million vs. 637).
The performance of the machine learning framework can also be seen in Fig.~\ref{fig:EulerianOT} and Fig.~\ref{fig:EulerianMFG}, which visualize characteristics and the push-forward and pull-back of $\rho_0$ and $\rho_1$, respectively.

\begin{figure*}
	\begin{center}
		\includegraphics[width=0.9\textwidth]{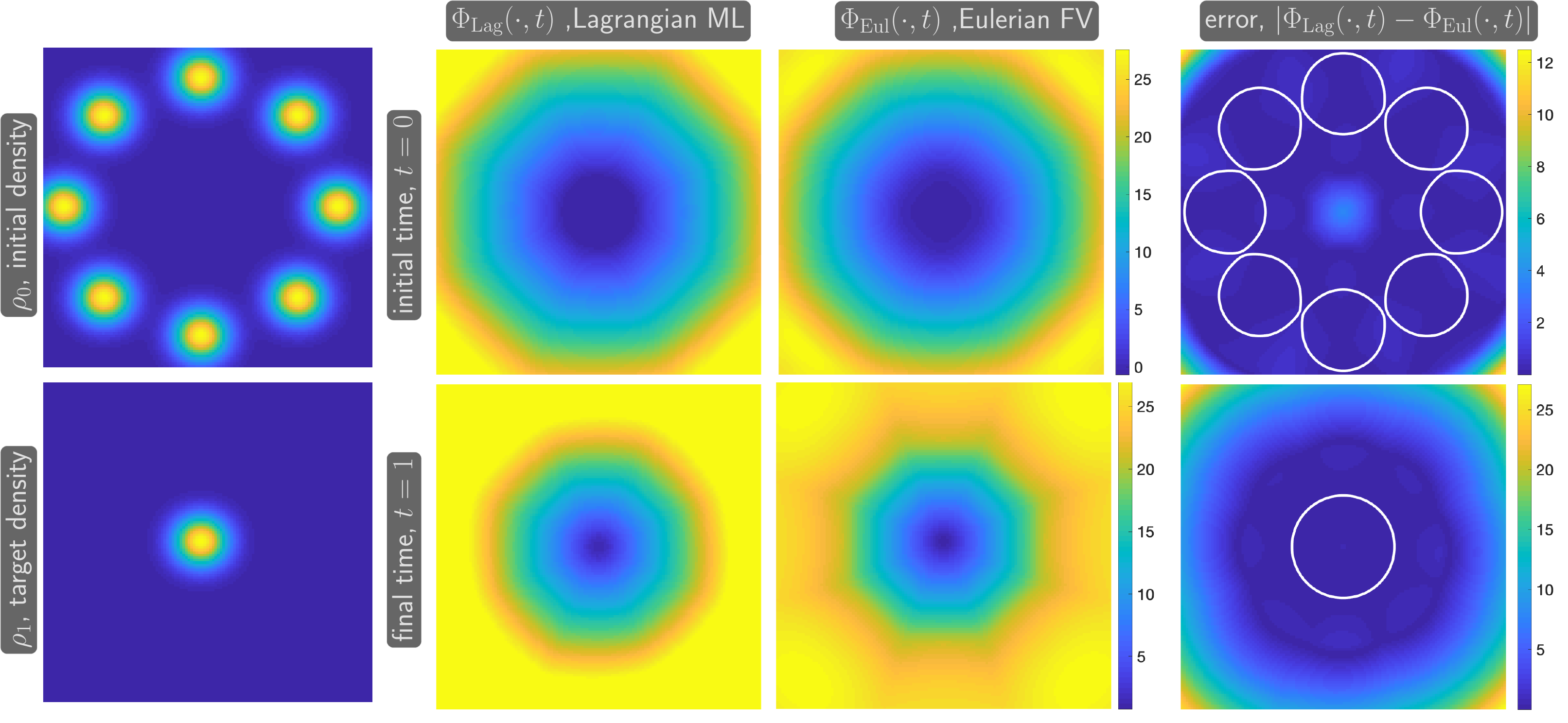}
	\end{center}
	\caption{Comparison of the potentials $\Phi_{\rm Lag}$, computed using our Lagrangian machine learning framework, and $\Phi_{\rm Eul}$, obtained from a provably convergent Eulerian method for the $d=2$-dimensional optimal transport example. 
    The left column shows the initial (first row) and target density (second row).
	The second column from the left shows $\Phi_{\rm Lag}$ at the initial time (top) and final time (bottom). 
	The third column from the left shows $\Phi_{\rm Eul}$ at the initial time (top) and final time (bottom). 
	The color axis are chosen equal in both plots to simplify the comparison. 
	The right column shows the absolute difference of the potentials obtained using both solvers and we illustrate the support $\rho(\cdot,t)$ by white contour lines. 
	As expected, the  estimated potentials match closely where mass in $\rho$ is concentrated and differ in other regions. 
	}
	\label{fig:EulerianOTPhi}
\end{figure*}

\begin{figure*}
	\begin{center}
		\includegraphics[width=0.9\textwidth]{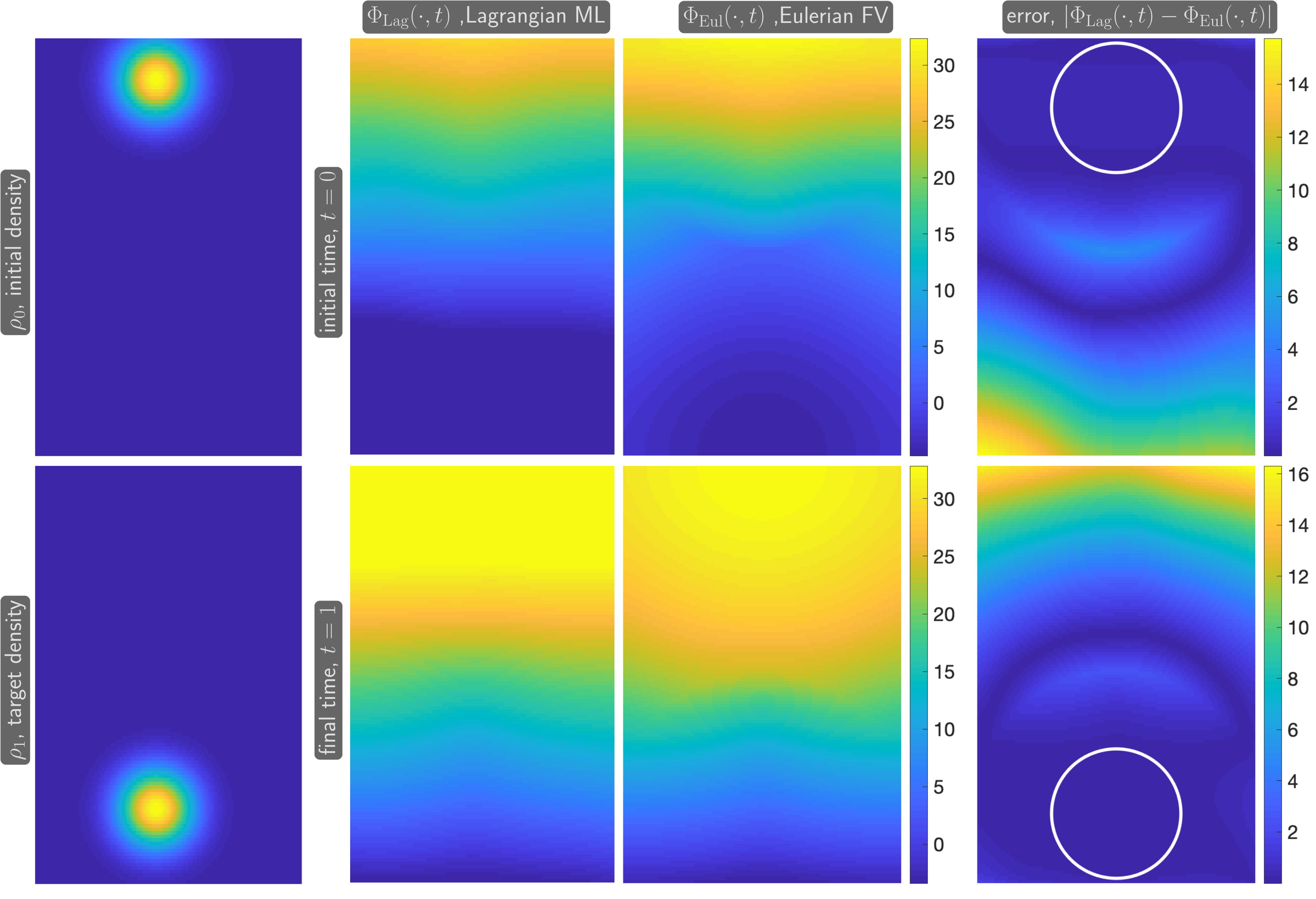}
	\end{center}
	\caption{Comparison of the potentials $\Phi_{\rm Lag}$, computed using our Lagrangian machine learning framework, and $\Phi_{\rm Eul}$, obtained from a provably convergent Eulerian method for the $d=2$-dimensional crowd motion problem. 
    The left column shows the initial (first row) and target density (second row).
	The second column from the left shows $\Phi_{\rm Lag}$ at the initial time (top) and final time (bottom). 
	The third column from the left shows $\Phi_{\rm Eul}$ at the initial time (top) and final time (bottom). 
	The color axis are chosen equal in both plots to simplify the comparison. 
	The right column shows the absolute difference of the potentials obtained using both solvers and we illustrate the support $\rho(\cdot,t)$ by white contour lines. 
	As expected, the  estimated potentials match closely where mass in $\rho$ is concentrated and differ in other regions. 
	}
	\label{fig:EulerianMFGPhi}
\end{figure*}

In Figs.~\ref{fig:EulerianOTPhi} and~\ref{fig:EulerianMFGPhi}, we provide a visual comparison of the potentials obtained using both methods for the optimal transport and crowd motion problem, respectively. 
In the Eulerian solver, we obtain the potential as the final Lagrange multiplier of the primal-dual Newton scheme.
In our machine learning framework, the potential can be computed by evaluating the trained network at the corresponding grid points.
We provide image visualizations of the potentials at the initial time and final time as well as their absolute difference.
As expected, the potentials agree in regions where $\rho(\cdot,t)$ is sufficiently large but the approximations can differ in other regions without affecting the cost functional.

\begin{table*}[t]
    \begin{center}
%

        \begin{tabular}{@{}|@{\;}c@{\;}|@{\;}c@{\;}|@{\;}c@{\;}|@{\;}c@{\;}|@{\;}c@{\;}|@{\;}c@{\;}|@{\;}c@{\;}|@{\;}c@{\;}|@{}} 
            \hline
			\multicolumn{8}{|c|}{Example 1: Optimal Transport} \\ \hline
                           & no. of parameters & $n_t$ & optimization                      &  $\mathcal{L}$  & $\mathcal{F}$& $\mathcal{G}$ & $\mathcal{J}_{\rm MFG}$ \\ \hline
            Eulerian, level 4   & 3,080,448    & -     &  Newton                           &  9.799e+00  & - &   8.625e-01   & 1.066e+01 (100.00\%)  \\
            Eulerian, level 3   & 376,960      & -     &  Newton                           &  9.764e+00  & - &   1.055e+00   & 1.082e+01 (101.47\%)  \\
            Lagrangian, ML      & 637          & 2     &  BFGS                             &  9.665e+00  & - &   1.059e+00   & 1.072e+01 (100.59\%) \\
            Lagrangian, ML      & 637          & 2     &  BFGS (no $\mathcal{C}_{\rm HJB}$)&  1.047e+01  & - &   3.759e+00   & 1.423e+01 (133.48\%) \\
            Lagrangian, ML      & 637          & 8     &  BFGS (no $\mathcal{C}_{\rm HJB}$)&  9.794e+00  & - &   8.344e-01   & 1.063e+01 (99.69\%) \\
            Lagrangian, ML      & 637          & 2     &  ADAM                             &  9.871e+00  & - &   8.294e-01   & 1.070e+01 (100.37\%) \\
			\hline
			\multicolumn{8}{|c|}{Example 2: Crowd Motion}      \\ \hline
			Eulerian, level 4   & 3,080,448    & -     &  Newton                           &  1.590e+01   & 2.274e+00 & 5.952e-01 & 1.877e+01 (100.00\%)\\
			Eulerian, level 3   &   376,960    & -     &  Newton                           &  1.590e+01   & 2.270e+00 & 6.729e-01 & 1.884e+01 (100.39\%)\\
            Lagrangian, ML      & 637          & 4     &  BFGS                             &  1.584e+01   & 2.275e+00 & 6.636e-01 & 1.878e+01 (100.08\%) \\
			\hline  \end{tabular}
    \end{center}
    \caption{Comparison of the optimal controls obtained using the Eulerian finite-volume approach and Lagrangian machine learning (ML) approach for the $d=2$ instances of the optimal transport and crowd motion problem. {\normalfont Legend: FV (finite volume), ML (machine learning), $n_t$ number of time integration steps for characteristics,  $\mathcal{L}$ (transport costs), $\mathcal{F}$ (running costs), $\mathcal{G} $ (terminal costs) , $\mathcal{J}_{\rm MFG} $ (objective functional) } }
    \label{tab:Eulerian}
\end{table*}


\end{appendix}

%% file: MachineLearningForMFG-arxiv.bbl
\begin{thebibliography}{10}

\bibitem{achdou13}
Y.~Achdou.
\newblock Finite difference methods for mean field games.
\newblock In {\em Hamilton-{J}acobi equations: approximations, numerical
  analysis and applications}, volume 2074 of {\em Lecture Notes in Math.},
  pages 1--47. Springer, Heidelberg, 2013.

\bibitem{moll14}
Y.~Achdou, F.~J. Buera, J.-M. Lasry, P.-L. Lions, and B.~Moll.
\newblock Partial differential equation models in macroeconomics.
\newblock {\em Philos. Trans. R. Soc. Lond. Ser. A Math. Phys. Eng. Sci.},
  372(2028):20130397, 19, 2014.

\bibitem{moll17}
Y.~Achdou, J.~Han, J.-M. Lasry, P.-L. Lions, and B.~Moll.
\newblock Income and wealth distribution in macroeconomics: A continuous-time
  approach.
\newblock Working Paper 23732, National Bureau of Economic Research, August
  2017.

\bibitem{achdou19}
Y.~Achdou and J.-M. Lasry.
\newblock Mean field games for modeling crowd motion.
\newblock In {\em Contributions to partial differential equations and
  applications}, volume~47 of {\em Comput. Methods Appl. Sci.}, pages 17--42.
  Springer, Cham, 2019.

\bibitem{achdou16}
Y.~Achdou and M.~Lauri\`ere.
\newblock Mean field type control with congestion ({II}): {A}n augmented
  {L}agrangian method.
\newblock {\em Appl. Math. Optim.}, 74(3):535--578, 2016.

\bibitem{gomes17}
N.~Almulla, R.~Ferreira, and D.~Gomes.
\newblock Two numerical approaches to stationary mean-field games.
\newblock {\em Dyn. Games Appl.}, 7(4):657--682, 2017.

\bibitem{Ambrosio2003}
L.~Ambrosio.
\newblock {Lecture notes on optimal transport problems}.
\newblock In {\em Mathematical aspects of evolving interfaces}, pages 1--52.
  Springer, Berlin, 2003.

\bibitem{ambrosio08}
L.~Ambrosio, N.~Gigli, and G.~Savar\'{e}.
\newblock {\em Gradient flows in metric spaces and in the space of probability
  measures}.
\newblock Lectures in Mathematics ETH Z\"{u}rich. Birkh\"{a}user Verlag, Basel,
  second edition, 2008.

\bibitem{aurell18}
A.~Aurell and B.~Djehiche.
\newblock Mean-field type modeling of nonlocal crowd aversion in pedestrian
  crowd dynamics.
\newblock {\em SIAM J. Control Optim.}, 56(1):434--455, 2018.

\bibitem{Bellman1957}
R.~Bellman.
\newblock {\em Dynamic programming}.
\newblock Princeton University Press, Princeton, N. J., 1957.

\bibitem{Bellman}
R.~Bellman.
\newblock {\em Introduction to Matrix Analysis (2Nd Ed.)}.
\newblock Society for Industrial and Applied Mathematics, Philadelphia, PA,
  USA, 1997.

\bibitem{BenamouBrenier2000}
J.~D. Benamou and Y.~Brenier.
\newblock {A computational fluid mechanics solution to the Monge-Kantorovich
  mass transfer problem}.
\newblock {\em Numerische Mathematik}, 2000.

\bibitem{bencarmarnen'18}
J.-D. Benamou, G.~Carlier, S.~Di~Marino, and L.~Nenna.
\newblock An entropy minimization approach to second-order variational
  mean-field games.
\newblock {\em Math. Models Methods Appl. Sci.}, 29(8):1553--1583, 2019.

\bibitem{bencarsan'17}
J.-D. Benamou, G.~Carlier, and F.~Santambrogio.
\newblock Variational mean field games.
\newblock In {\em Active particles. {V}ol. 1. {A}dvances in theory, models, and
  applications}, Model. Simul. Sci. Eng. Technol., pages 141--171.
  Birkh\"auser/Springer, Cham, 2017.

\bibitem{bensoussan2013}
A.~Bensoussan, J.~Frehse, and P.~Yam.
\newblock {\em Mean field games and mean field type control theory}.
\newblock SpringerBriefs in Mathematics. Springer, New York, 2013.

\bibitem{Bezanson:2017gd}
J.~Bezanson, A.~Edelman, S.~Karpinski, and V.~B. Shah.
\newblock {Julia: A Fresh Approach to Numerical Computing}.
\newblock {\em SIAM review}, 59(1):65--98, Jan. 2017.

\bibitem{bottou2016optimization}
L.~Bottou, F.~E. Curtis, and J.~Nocedal.
\newblock {Optimization methods for large-scale machine learning}.
\newblock {\em SIAM Review}, 60(2):223--311, 2018.

\bibitem{silva19}
L.~Brice\~{n}o Arias, D.~Kalise, Z.~Kobeissi, M.~Lauri\`ere,
  A.~Mateos~Gonz\'{a}lez, and F.~J. Silva.
\newblock On the implementation of a primal-dual algorithm for second order
  time-dependent mean field games with local couplings.
\newblock In {\em C{EMRACS} 2017---numerical methods for stochastic models:
  control, uncertainty quantification, mean-field}, volume~65 of {\em ESAIM
  Proc. Surveys}, pages 330--348. EDP Sci., Les Ulis, 2019.

\bibitem{wolfram14}
M.~Burger, M.~Di~Francesco, P.~A. Markowich, and M.-T. Wolfram.
\newblock Mean field games with nonlinear mobilities in pedestrian dynamics.
\newblock {\em Discrete Contin. Dyn. Syst. Ser. B}, 19(5):1311--1333, 2014.

\bibitem{carda19}
P.~Cardaliaguet, F.~Delarue, J.-M. Lasry, and P.-L. Lions.
\newblock {\em The master equation and the convergence problem in mean field
  games}, volume 201 of {\em Annals of Mathematics Studies}.
\newblock Princeton University Press, Princeton, NJ, 2019.

\bibitem{lehalle18}
P.~Cardaliaguet and C.-A. Lehalle.
\newblock Mean field game of controls and an application to trade crowding.
\newblock {\em Math. Financ. Econ.}, 12(3):335--363, 2018.

\bibitem{silva15}
E.~Carlini and F.~J. Silva.
\newblock A semi-{L}agrangian scheme for a degenerate second order mean field
  game system.
\newblock {\em Discrete Contin. Dyn. Syst.}, 35(9):4269--4292, 2015.

\bibitem{delarue18a}
R.~Carmona and F.~Delarue.
\newblock {\em Probabilistic theory of mean field games with applications.
  {I}}, volume~83 of {\em Probability Theory and Stochastic Modelling}.
\newblock Springer, Cham, 2018.
\newblock Mean field FBSDEs, control, and games.

\bibitem{delarue18b}
R.~Carmona and F.~Delarue.
\newblock {\em Probabilistic theory of mean field games with applications.
  {II}}, volume~84 of {\em Probability Theory and Stochastic Modelling}.
\newblock Springer, Cham, 2018.
\newblock Mean field games with common noise and master equations.

\bibitem{CarmonaLauriere2019_convergencea}
R.~Carmona and M.~Lauri{\`e}re.
\newblock Convergence {{Analysis}} of {{Machine Learning Algorithms}} for the
  {{Numerical Solution}} of {{Mean Field Control}} and {{Games}}: {{I}} --
  {{The Ergodic Case}}.
\newblock {\em arXiv:1907.05980 [cs, math]}, 2019.

\bibitem{CarmonaLauriere2019_convergence}
R.~Carmona and M.~Lauri{\`e}re.
\newblock Convergence {{Analysis}} of {{Machine Learning Algorithms}} for the
  {{Numerical Solution}} of {{Mean Field Control}} and {{Games}}: {{II}} --
  {{The Finite Horizon Case}}.
\newblock {\em arXiv:1908.01613 [cs, math]}, 2019.

\bibitem{jaimungal19}
P.~Casgrain and S.~Jaimungal.
\newblock Algorithmic trading in competitive markets with mean field games.
\newblock {\em SIAM News}, 52(2), 2019.

\bibitem{cirant19}
A.~Cesaroni and M.~Cirant.
\newblock Introduction to variational methods for viscous ergodic mean-field
  games with local coupling.
\newblock In {\em Contemporary research in elliptic {PDE}s and related topics},
  volume~33 of {\em Springer INdAM Ser.}, pages 221--246. Springer, Cham, 2019.

\bibitem{Chang:2017te}
B.~Chang, L.~Meng, E.~Haber, F.~Tung, and D.~Begert.
\newblock Multi-level residual networks from dynamical systems view.
\newblock In {\em International Conference on Learning Representations}, 2018.

\bibitem{Chow2019}
Y.~T. Chow, W.~Li, S.~Osher, and W.~Yin.
\newblock Algorithm for {H}amilton--{J}acobi equations in density space via a
  generalized {H}opf formula.
\newblock {\em J Sci Comput}, 80(2):1195--1239, Aug 2019.

\bibitem{paola19}
A.~{De Paola}, V.~{Trovato}, D.~{Angeli}, and G.~{Strbac}.
\newblock A mean field game approach for distributed control of thermostatic
  loads acting in simultaneous energy-frequency response markets.
\newblock {\em IEEE Transactions on Smart Grid}, 10(6):5987--5999, Nov 2019.

\bibitem{E:2017kz}
W.~E.
\newblock {A Proposal on Machine Learning via Dynamical Systems}.
\newblock {\em Communications in Mathematics and Statistics}, 5(1):1--11, Mar.
  2017.

\bibitem{E2016}
W.~E, J.~Han, and A.~Jentzen.
\newblock Deep learning-based numerical methods for high-dimensional parabolic
  partial differential equations and backward stochastic differential
  equations.
\newblock {\em Communications in Mathematics and Statistics}, 5:349--380, 2017.

\bibitem{EHanLi2018_meanfield}
W.~E, J.~Han, and Q.~Li.
\newblock A {{Mean}}-{{Field Optimal Control Formulation}} of {{Deep
  Learning}}.
\newblock {\em arXiv:1807.01083 [cs, math]}, 2018.

\bibitem{nurbe18}
D.~Evangelista, R.~Ferreira, D.~A. Gomes, L.~Nurbekyan, and V.~Voskanyan.
\newblock First-order, stationary mean-field games with congestion.
\newblock {\em Nonlinear Anal.}, 173:37--74, 2018.

\bibitem{Evans1997}
L.~C. Evans.
\newblock {Partial differential equations and Monge-Kantorovich mass transfer}.
\newblock {\em Current developments in mathematics}, 1997.

\bibitem{evans10}
L.~C. Evans.
\newblock {\em Partial differential equations}, volume~19 of {\em Graduate
  Studies in Mathematics}.
\newblock American Mathematical Society, Providence, RI, second edition, 2010.

\bibitem{caines17}
D.~{Firoozi} and P.~E. {Caines}.
\newblock An optimal execution problem in finance targeting the market trading
  speed: An mfg formulation.
\newblock In {\em 2017 IEEE 56th Annual Conference on Decision and Control
  (CDC)}, pages 7--14, Dec 2017.

\bibitem{soner2006}
W.~H. Fleming and H.~M. Soner.
\newblock {\em Controlled {M}arkov processes and viscosity solutions},
  volume~25 of {\em Stochastic Modelling and Applied Probability}.
\newblock Springer, New York, 2nd edition, 2006.

\bibitem{gangbo15}
W.~Gangbo and A.~\'{S}wi\k{e}ch.
\newblock Existence of a solution to an equation arising from the theory of
  mean field games.
\newblock {\em J. Differential Equations}, 259(11):6573--6643, 2015.

\bibitem{Gomes:2015th}
D.~A. Gomes, L.~Nurbekyan, and E.~A. Pimentel.
\newblock {\em Economic models and mean-field games theory}.
\newblock IMPA Mathematical Publications. Instituto Nacional de Matem\'atica
  Pura e Aplicada (IMPA), Rio de Janeiro, 2015.

\bibitem{gomes_book16}
D.~A. Gomes, E.~A. Pimentel, and V.~Voskanyan.
\newblock {\em Regularity theory for mean-field game systems}.
\newblock SpringerBriefs in Mathematics. Springer, [Cham], 2016.

\bibitem{gomessaude'18}
D.~A. Gomes and J.~Sa\'{u}de.
\newblock A mean-field game approach to price formation in electricity markets.
\newblock {\em Preprint}, 2018.
\newblock arXiv:1807.07088 [math.AP].

\bibitem{gueantlasrylions11}
O.~Gu\'{e}ant, J.-M. Lasry, and P.-L. Lions.
\newblock Mean field games and applications.
\newblock In {\em Paris-{P}rinceton {L}ectures on {M}athematical {F}inance
  2010}, volume 2003 of {\em Lecture Notes in Math.}, pages 205--266. Springer,
  Berlin, 2011.

\bibitem{GuntherEtAl2018}
S.~G{\"u}nther, L.~Ruthotto, J.~B. Schroder, E.~C. Cyr, and N.~R. Gauger.
\newblock {Layer-Parallel Training of Deep Residual Neural Networks}.
\newblock {\em Journal on Mathematical Imaging and Vision}, math.OC:1--23,
  2019.

\bibitem{HaberHoresh2015}
E.~Haber and R.~Horesh.
\newblock {A Multilevel Method for the Solution of Time Dependent Optimal
  Transport}.
\newblock {\em Numer. Math. Theory Methods Appl.}, 8(01):97--111, Mar. 2015.

\bibitem{HaberRuthotto2017}
E.~Haber and L.~Ruthotto.
\newblock {Stable architectures for deep neural networks}.
\newblock {\em Inverse Problems}, 34(1):1--22, 2017.

\bibitem{Han:2016ku}
J.~Han and W.~E.
\newblock {Deep Learning Approximation for Stochastic Control Problems}.
\newblock {\em arXiv preprint arXiv:1611.07422}, Nov. 2016.

\bibitem{HanEtAl2017}
J.~Han, A.~Jentzen, and W.~E.
\newblock Solving high-dimensional partial differential equations using deep
  learning.
\newblock {\em Proceedings of the National Academy of Sciences},
  115(34):8505--8510, 2018.

\bibitem{He:2016tt}
K.~He, X.~Zhang, S.~Ren, and J.~Sun.
\newblock {Deep Residual Learning for Image Recognition}.
\newblock pages 770--778, 2016.

\bibitem{HCM07}
M.~Huang, P.~E. Caines, and R.~P. Malham\'e.
\newblock Large-population cost-coupled {LQG} problems with nonuniform agents:
  individual-mass behavior and decentralized {$\epsilon$}-{N}ash equilibria.
\newblock {\em IEEE Trans. Automat. Control}, 52(9):1560--1571, 2007.

\bibitem{HCM06}
M.~Huang, R.~P. Malham\'e, and P.~E. Caines.
\newblock Large population stochastic dynamic games: closed-loop
  {M}c{K}ean-{V}lasov systems and the {N}ash certainty equivalence principle.
\newblock {\em Commun. Inf. Syst.}, 6(3):221--251, 2006.

\bibitem{innes:2018}
M.~Innes.
\newblock Flux: Elegant machine learning with julia.
\newblock {\em Journal of Open Source Software}, 2018.

\bibitem{matt'18}
M.~Jacobs and F.~L\'{e}ger.
\newblock A fast approach to optimal transport: the back-and-forth method.
\newblock {\em Preprint}, 2019.
\newblock arXiv:1905.12154 [math.OC].

\bibitem{JacobsLegerLiOsher2019_solving}
M.~Jacobs, F.~L{\'e}ger, W.~Li, and S.~Osher.
\newblock Solving {{Large}}-{{Scale Optimization Problems}} with a
  {{Convergence Rate Independent}} of {{Grid Size}}.
\newblock {\em SIAM J. Numer. Anal.}, 2019.

\bibitem{Kantorovich2006}
L.~V. Kantorovich.
\newblock {On a Problem of Monge}.
\newblock {\em J. Math. Sci.}, 133(4):1383--1383, Mar. 2006.

\bibitem{kingma2014adam}
D.~P. Kingma and J.~Ba.
\newblock Adam: A method for stochastic optimization.
\newblock {\em arXiv preprint arXiv:1412.6980}, 2014.

\bibitem{kizikale19}
A.~C. Kizilkale, R.~Salhab, and R.~P. Malhamé.
\newblock An integral control formulation of mean field game based large scale
  coordination of loads in smart grids.
\newblock {\em Automatica}, 100:312 -- 322, 2019.

\bibitem{lehalle16}
A.~Lachapelle, J.-M. Lasry, C.-A. Lehalle, and P.-L. Lions.
\newblock Efficiency of the price formation process in presence of high
  frequency participants: a mean field game analysis.
\newblock {\em Math. Financ. Econ.}, 10(3):223--262, 2016.

\bibitem{wolfram11}
A.~Lachapelle and M.-T. Wolfram.
\newblock On a mean field game approach modeling congestion and aversion in
  pedestrian crowds.
\newblock {\em Transp. Res. Part B: Methodol.}, 45(10):1572 -- 1589, 2011.

\bibitem{LasryLions06a}
J.-M. Lasry and P.-L. Lions.
\newblock Jeux \`a champ moyen. {I}. {L}e cas stationnaire.
\newblock {\em C. R. Math. Acad. Sci. Paris}, 343(9):619--625, 2006.

\bibitem{LasryLions06b}
J.-M. Lasry and P.-L. Lions.
\newblock Jeux \`a champ moyen. {II}. {H}orizon fini et contr\^{o}le optimal.
\newblock {\em C. R. Math. Acad. Sci. Paris}, 343(10):679--684, 2006.

\bibitem{LasryLions2007}
J.-M. Lasry and P.-L. Lions.
\newblock Mean field games.
\newblock {\em Jpn. J. Math.}, 2(1):229--260, 2007.

\bibitem{Li:2017ta}
H.~Li, Z.~Xu, G.~Taylor, and T.~Goldstein.
\newblock {Visualizing the loss landscape of neural nets}.
\newblock In {\em papers.nips.cc}, 2018.

\bibitem{Li:2017wr}
Q.~Li, L.~Chen, C.~Tai, and W.~E.
\newblock {Maximum Principle Based Algorithms for Deep Learning}.
\newblock {\em arXiv preprint arXiv:1710.09513}, Oct. 2017.

\bibitem{Li2018ConstrainedDO}
W.~Li and S.~Osher.
\newblock Constrained dynamical optimal transport and its lagrangian
  formulation.
\newblock {\em arXiv preprint arXiv:1807.00937}, 2018.

\bibitem{LiRyuOsherYinGangbo2017_parallel}
W.~Li, E.~K. Ryu, S.~Osher, W.~Yin, and W.~Gangbo.
\newblock A {{Parallel Method}} for {{Earth Mover}}'s {{Distance}}.
\newblock {\em Journal of Scientific Computing}, pages 1--16, 2017.

\bibitem{Lin:2019ui}
J.~Lin, K.~Lensink, and E.~Haber.
\newblock {Fluid Flow Mass Transport for Generative Networks}.
\newblock {\em arXiv preprint ariXiv:1910.01694}, Oct. 2019.

\bibitem{MangRuthotto2017}
A.~Mang and L.~Ruthotto.
\newblock {A Lagrangian Gauss-Newton-Krylov Solver for Mass- and
  Intensity-Preserving Diffeomorphic Image Registration}.
\newblock {\em SIAM J. Sci. Comput.}, 39(5):B860--B885, 2017.

\bibitem{Modersitzki2009}
J.~Modersitzki.
\newblock {\em {FAIR: flexible algorithms for image registration}}, volume~6 of
  {\em Fundamentals of Algorithms}.
\newblock Society for Industrial and Applied Mathematics (SIAM), Philadelphia,
  PA, 2009.

\bibitem{NemirovskiEtAl2009}
A.~Nemirovski, A.~Juditsky, G.~Lan, and A.~Shapiro.
\newblock {Robust Stochastic Approximation Approach to Stochastic Programming}.
\newblock {\em SIAM Journal on Optimization}, 19(4):1574--1609, Jan. 2009.

\bibitem{NocedalWright2006}
J.~Nocedal and S.~Wright.
\newblock {\em {Numerical Optimization}}.
\newblock Springer Series in Operations Research and Financial Engineering.
  Springer Science {\&} Business Media, New York, Dec. 2006.

\bibitem{nursaude18}
L.~Nurbekyan and J.~Sa\'{u}de.
\newblock Fourier approximation methods for first-order nonlocal mean-field
  games.
\newblock {\em Port. Math.}, 75(3-4):367--396, 2018.

\bibitem{PeyreCuturi2018_computationalb}
G.~Peyré and M.~Cuturi.
\newblock Computational optimal transport.
\newblock {\em Foundations and Trends in Machine Learning}, 11 (5-6):355--602,
  2019.

\bibitem{Rezende:2015vu}
D.~J. Rezende and S.~Mohamed.
\newblock {Variational Inference with Normalizing Flows}.
\newblock In {\em ICML'15: Proceedings of the 32Nd International Conference on
  International Conference on Machine Learning - Volume 37}. JMLR.org, 2015.

\bibitem{Robbins:1951ko}
H.~Robbins and S.~Monro.
\newblock {A Stochastic Approximation Method}.
\newblock {\em The annals of mathematical statistics}, 22(3):400--407, 1951.

\bibitem{Sirignano:ik}
J.~Sirignano and K.~Spiliopoulos.
\newblock {DGM: A deep learning algorithm for solving partial differential
  equations}.
\newblock {\em J. Comput. Phys.}, (375):1339--1364, 2018.

\bibitem{Villani2003}
C.~Villani.
\newblock {\em {Topics in Optimal Transportation}}.
\newblock American Mathematical Soc., 2003.

\bibitem{villani2008optimal}
C.~Villani.
\newblock {\em Optimal transport: old and new}, volume 338.
\newblock Springer Science, 2008.

\bibitem{welch19}
P.~M. Welch, K.~{\O}. Rasmussen, and C.~F. Welch.
\newblock Describing nonequilibrium soft matter with mean field game theory.
\newblock {\em The Journal of Chemical Physics}, 150(17):174905, 2019.

\bibitem{YangYeTrivediXuZha2017_learning}
J.~Yang, X.~Ye, R.~Trivedi, H.~Xu, and H.~Zha.
\newblock Deep mean field games for learning optimal behavior policy of large
  populations.
\newblock In {\em International Conference on Learning Representations}, 2018.

\bibitem{YangEtAl2019}
L.~Yang and G.~E. Karniadakis.
\newblock {Potential Flow Generator with L$_{2}$ Optimal Transport Regularity
  for Generative Models}.
\newblock {\em arXiv.org}, Aug. 2019.

\bibitem{ZhangEtAl2018}
L.~Zhang, W.~E, and L.~Wang.
\newblock {Monge-Amp{\`e}re Flow for Generative Modeling}.
\newblock {\em arXiv.org}, Sept. 2018.

\end{thebibliography}
